\documentclass[letterpaper]{article} 
\usepackage{aaai25}  
\usepackage{times}  
\usepackage{helvet}  
\usepackage{courier}  
\usepackage[hyphens]{url}  
\usepackage{graphicx} 
\usepackage{natbib}  
\usepackage{caption} 
\usepackage{algorithm}
\usepackage{algorithmic}

\usepackage{newfloat}
\usepackage{listings}

\DeclareCaptionStyle{ruled}{labelfont=normalfont,labelsep=colon,strut=off} 
\floatstyle{ruled}
\newfloat{listing}{tb}{lst}{}
\floatname{listing}{Listing}
\title{VFG-SSC: Semi-supervised 3D Semantic Scene Completion \\with 2D Vision Foundation
Model Guidance}
\author{
    Duc-Hai Pham\textsuperscript{\rm 1},
    Duc-Dung Nguyen\textsuperscript{\rm 2},
    Anh Pham\textsuperscript{\rm 2},
    Tuan Ho\textsuperscript{\rm 1}, 
    Phong Nguyen\textsuperscript{\rm 1}, \\
    Khoi Nguyen\textsuperscript{\rm 1},
    Rang Nguyen\textsuperscript{\rm 1}
}

\newcommand{\myheading}[1]{\vspace{.75ex}\noindent \textbf{#1}}

\usepackage{amsfonts}
\usepackage{amsmath}
\usepackage{booktabs}
\usepackage{multirow}
\usepackage{xcolor, colortbl}
\usepackage[capitalize]{cleveref}
\crefname{section}{Sec.}{Secs.}
\Crefname{section}{Section}{Sections}
\Crefname{table}{Table}{Tables}
\crefname{table}{Tab.}{Tabs.}
 
\definecolor{mydarkblue}{rgb}{0,0.08,1}
\definecolor{mydarkgreen}{rgb}{0.02,0.6,0.02}
\definecolor{myred}{rgb}{1.0,0.0,0.0}
\definecolor{myred2}{rgb}{0.7,0.1,0.1}
\definecolor{mydarkblue2}{rgb}{0.05,0.1,0.7}
\definecolor{mypurple}{rgb}{111,0,255}
\definecolor{mypurple2}{rgb}{111,0,111}

\definecolor{ceiling}{RGB}{214,  38, 40}   %
\definecolor{floor}{RGB}{43, 160, 4}     %
\definecolor{wall}{RGB}{158, 216, 229}  %
\definecolor{window}{RGB}{114, 158, 206}  %
\definecolor{chair}{RGB}{204, 204, 91}   %
\definecolor{bed}{RGB}{255, 186, 119}  %
\definecolor{sofa}{RGB}{147, 102, 188}  %
\definecolor{table}{RGB}{30, 119, 181}   %
\definecolor{tvs}{RGB}{160, 188, 33}   %
\definecolor{furniture}{RGB}{255, 127, 12}  %
\definecolor{objects}{RGB}{196, 175, 214} %

\definecolor{car}{rgb}{0.39215686, 0.58823529, 0.96078431}
\definecolor{bicycle}{rgb}{0.39215686, 0.90196078, 0.96078431}
\definecolor{motorcycle}{rgb}{0.11764706, 0.23529412, 0.58823529}
\definecolor{truck}{rgb}{0.31372549, 0.11764706, 0.70588235}
\definecolor{other-vehicle}{rgb}{0.39215686, 0.31372549, 0.98039216}
\definecolor{person}{rgb}{1.        , 0.11764706, 0.11764706}
\definecolor{bicyclist}{rgb}{1.        , 0.15686275, 0.78431373}
\definecolor{motorcyclist}{rgb}{0.58823529, 0.11764706, 0.35294118}
\definecolor{road}{rgb}{1.        , 0.        , 1.        }
    \definecolor{parking}{rgb}{1.        , 0.58823529, 1.        }
\definecolor{sidewalk}{rgb}{0.29411765, 0.        , 0.29411765}
\definecolor{other-ground}{rgb}{0.68627451, 0.        , 0.29411765}
\definecolor{building}{rgb}{1.        , 0.78431373, 0.        }
\definecolor{fence}{rgb}{1.        , 0.47058824, 0.19607843}
\definecolor{vegetation}{rgb}{0.        , 0.68627451, 0.        }
\definecolor{trunk}{rgb}{0.52941176, 0.23529412, 0.        }
\definecolor{terrain}{rgb}{0.58823529, 0.94117647, 0.31372549}
\definecolor{pole}{rgb}{1.        , 0.94117647, 0.58823529}
\definecolor{traffic-sign}{rgb}{1.        , 0.        , 0.    }

\def\Approach{Semi-SSC}

\usepackage{amssymb}%
\usepackage{pifont}
\newcommand{\cmark}{\ding{51}}%

\affiliations{
    \textsuperscript{\rm 1}VinAI Research, Vietnam \\
    \textsuperscript{\rm 2}AITech Lab., Ho Chi Minh City University of Technology, VNU-HCM, Vietnam\\

    \{v.haipd13, v.tuanhl25, v.phongnh31, v.khoindm, v.rangnhm\}@vinai.io, \{anhpham, nddung\}@hcmut.edu.vn\\

}

\usepackage{bibentry}

\begin{document}

\maketitle

\begin{abstract}
Accurate prediction of 3D semantic occupancy from 2D visual images is crucial for enabling autonomous agents to understand their surroundings for planning and navigation. State-of-the-art methods typically rely on fully supervised approaches, requiring large labeled datasets obtained through expensive LiDAR sensors and meticulous voxel-wise annotation by human experts. The resource-intensive nature of this annotation process significantly limits the scalability and application of these methods. To address this challenge, we propose a novel semi-supervised framework that reduces reliance on densely annotated data. Our approach leverages 2D foundation models to extract essential 3D scene geometry and semantic cues, enabling a more efficient training process. The proposed framework has two key advantages: (1) \textbf{Generalizability}, as it is compatible with various 3D semantic scene completion methods, including 2D-3D lifting and 3D-2D transformer techniques; and (2) \textbf{Effectiveness}, as demonstrated by experiments on the SemanticKITTI and NYUv2 datasets, where our method achieves up to 85\% of the fully supervised performance using only 10\% of the labeled data. This approach not only reduces the cost of data annotation but also highlights its potential for broader adoption in vision-based systems for 3D semantic occupancy prediction.
\end{abstract}

\section{Introduction}
\label{sec:intro}

Vision-based 3D semantic scene completion (SSC) is a crucial vision task with numerous applications in both indoor and outdoor scenarios \cite{SSCNet, liu2018see, roldao2020lmscnet, wu2020scfusion, li2020anisotropic, dourado2021edgenet, huang2023tri, MonoScene, SceneAsOcc, yao2023ndc, VoxFormer, OccFormer}. Given an input image, the goal is to predict a complete 3D voxel representation of volumetric occupancy and semantic labels for a scene. This task is challenging due to the absence of 3D information in 2D images. Most SSC methods \cite{SSCNet, VoxFormer, OccFormer} address this challenge through fully supervised learning, which relies on dense 3D SSC annotations. However, obtaining these annotations requires deploying vehicles equipped with LiDAR sensors and performing meticulous voxel-level human labeling, which severely limits the scalability and deployment of many SSC systems. An alternative approach to reduce the annotation costs is self-supervised learning~\cite{cao2023scenerf, huang2023selfocc, zhang2023occnerf}, which only requires unlabeled images for training. However, these methods often produce suboptimal results, with a significant performance gap compared to fully supervised approaches.

\begin{figure}[tb]
  \centering
  \includegraphics[width=\columnwidth]{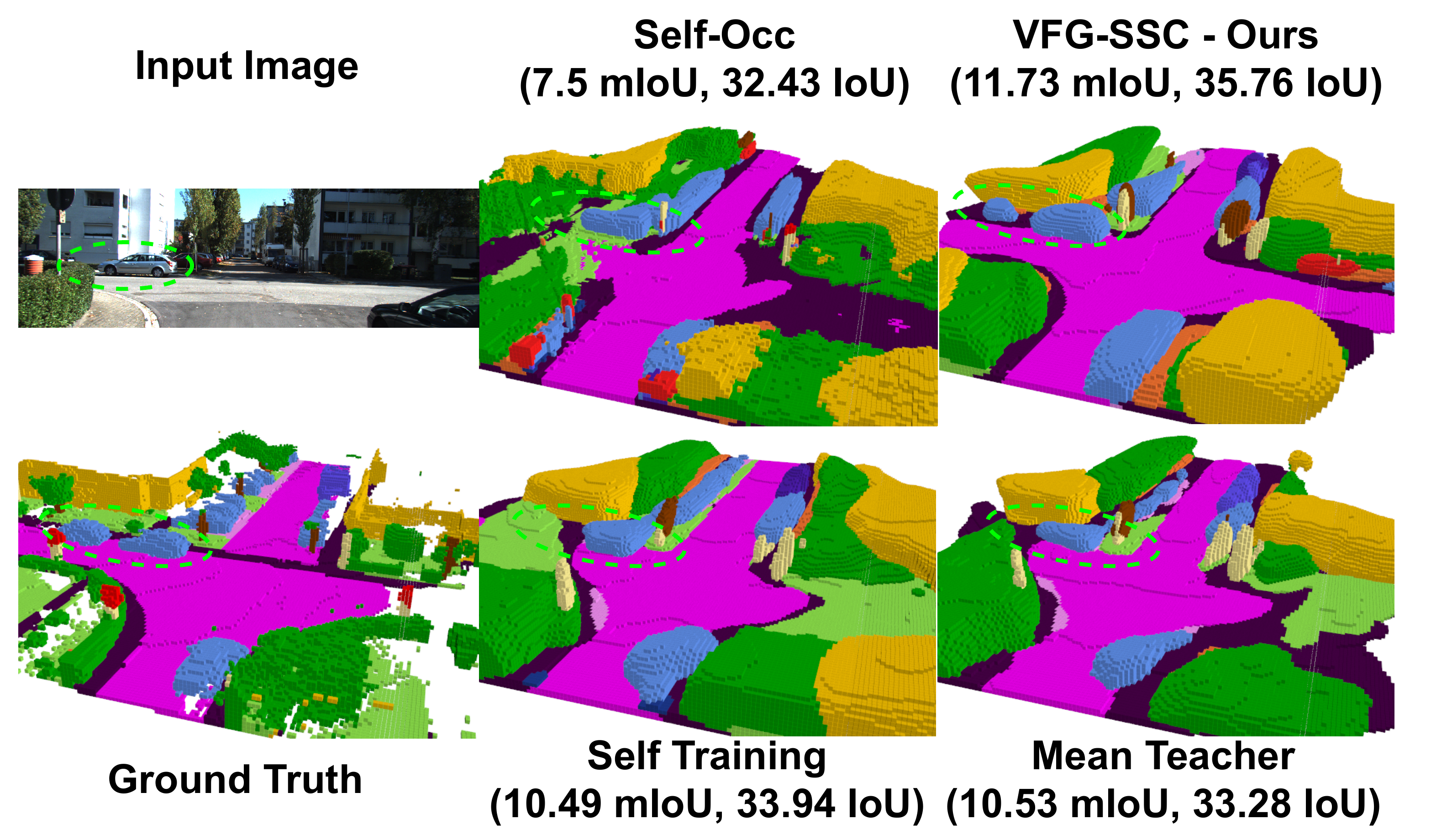}
  \caption{\textbf{Visual comparison on SemanticKITTI}: VFG-SSC surparses self-supervised methods like SelfOcc by 4\% mIoU and outperforms other semi-supervised methods: Self-Training and Mean Teacher by 1.5\% mIoU.}
  \label{fig:overview}
\end{figure}

To address these limitations, this paper introduces the problem of semi-supervised vision-based 3D semantic scene completion (\Approach). In this setup, a small percentage of images are labeled with 3D semantic occupancy, while the majority remain unlabeled. To overcome the challenge of limited labeled data, semi-supervised learning approaches like \textit{Self-Training} \cite{Self-Training} and \textit{Mean-Teacher} \cite{MeanTeacher} can be employed. In Self-Training (ST), a fully supervised SSC network first learns from labeled data and then predicts pseudo-labels for unlabeled data. These pseudo-labels are used to retrain the network, leveraging the unlabeled data. In contrast, the Mean-Teacher (MT) approach involves a more stable Teacher network, derived from the student network's Exponential Moving Average (EMA), which predicts pseudo-labels for unlabeled images. These pseudo-labels are combined with labeled data to continually train the student network. While there are advanced semi-supervised learning techniques for 3D, such as LaserMix \cite{LaserMix}, CutMix \cite{CutMix}, and ClassMix \cite{ClassMix}, these methods rely on a one-to-one correspondence between input and output or homogeneous data. This approach does not apply to Semi-SSC, where the input consists of 2D images and the output is in 3D voxels.

On the other hand, recent vision foundation models (VFMs) have shown the ability to robustly extract metric depth \cite{piccinelli2024unidepth} and semantics \cite{OpenSEED} from unlabeled images. However, these capabilities are not yet leveraged in SSC. To bridge this gap, we introduce VFG-SSC, a method that leverages 2D VFMs to generate 3D priors from unlabeled images, thereby improving the quality of pseudo-labels during training. Adapting 2D VFM knowledge to 3D presents challenges such as depth scale ambiguity and feature misalignment from segmentation models. To address these, we developed a novel attention-based enhancement module that (1) globally refines features and (2) maintains linear complexity, enabling efficient processing of large 3D datasets. Futhermore, we propose a technique to accumulate VFM information over image sequences, reducing noise and biases compared to using a single image. Our method can serve as a universal plugin for any vision-based SSC method, enabling a semi-supervised alternative for fully supervised approaches.

We evaluate our approach on the outdoor SemanticKITTI \cite{SemanticKITTI} and indoor NYUv2 \cite{silberman2012indoor} datasets. Our results demonstrate strong performance across both settings, achieving 85\% of the accuracy of fully supervised methods while utilizing only 10\% of the 3D semantic occupancy annotations.

In summary, our contributions are as follows:

\begin{itemize}
    \item We introduce a new semi-supervised 3D semantic scene completion (\Approach) framework to reduce the high cost of 3D voxel-wise annotation.
    \item We propose VFG-SSC, a novel approach that leverages 3D cues from 2D vision foundation models for depth and segmentation.
    \item We design a new enhancement module to effectively combine 3D cues with 3D features, seamlessly integrating with existing SSC methods.
    \item Our pipeline achieves 85\% of fully supervised performance using just 10\% of the labeled data.
\end{itemize}

\section{Related Work}
\label{sec:related_work}
This section reviews literature on fully, semi, and self-supervised 3D semantic scene completion, followed by an overview of the 2D depth and semantic segmentation foundation models used in our framework.

\myheading{3D Semantic Scene Completion (SSC).} The primary goal of 3D semantic scene completion (SSC) or 3D semantic occupancy prediction is to jointly infer the geometry and semantics of a scene. Initially introduced in SSCNet \cite{SSCNet} for indoor scenes like NYUv2 \cite{silberman2012indoor}, the task has since been extended to outdoor environments such as SemanticKITTI \cite{SemanticKITTI} \cite{roldao2020lmscnet, li2020anisotropic, yan2021sparse, S3CNet, SSC-RS, SCPNet, jia2023occupancydetr}. 
\textbf{Vision-based SSC} methods, which are cost-effective and easily deployable, have gained traction. MonoScene \cite{MonoScene} was the first to use only RGB images for SSC by projecting 3D voxels onto 2D images and processing them with a 3D UNet to predict 3D semantic occupancy. VoxFormer \cite{VoxFormer} improves upon this by leveraging depth estimation priors to focus cross-attention on voxels near object surfaces. OccFormer \cite{OccFormer} further enhances 3D feature encoding using windowed attention on local and global transformer pathways and employs a mask decoder, similar to Mask2Former \cite{Mask2Former}, to predict 3D SSC. However, these methods still rely on expensive voxel-wise annotations, limiting their practical application.

\myheading{Self-supervised SSC.} One way to reduce the annotation cost of SSC is through self-supervised learning, which leverages video sequences along with predicted depth and semantic segmentation as pseudo-labels. Methods like SceneRF \cite{cao2023scenerf}, SelfOcc \cite{huang2023selfocc}, and OccNeRF \cite{zhang2023occnerf} use volumetric rendering on predicted 3D semantic occupancy to generate RGB images, depth maps, and segmentation maps, allowing for loss computation with pseudo-labels. Although self-supervised SSC requires less supervision than our VFG-SSC approach, its performance is significantly lower than that of semi-supervised and fully-supervised methods, making it unsuitable for practical applications.


\myheading{Semi-supervised learning for 2D and 3D LiDAR segmentation.} Semi-supervised image segmentation methods \cite{Semi-CPS, PSMT, SSL-ELN, U2PL, UniMatch, SemiVL} have demonstrated strong performance on autonomous datasets by leveraging two key strategies: consistency regularization and entropy minimization. Consistency regularization employs augmentations to ensure stable predictions under perturbed inputs, while entropy minimization reduces uncertainty in the classification of unlabeled data, leading to more confident predictions. In 3D LiDAR segmentation, LaserMix \cite{LaserMix} enforces consistency through mixing schemes, while methods like 3DIoUMatch \cite{3DIoUMatch} and DetMatch \cite{DetMatch} use multi-sensor data to generate high-quality pseudo-labels for object detection. However, these methods rely on a homogeneous input-output relationship. In contrast, \Approach presents a greater challenge as it takes 2D images as input and produces 3D semantic occupancy as output. Consequently, directly applying the above methods to Semi-SSC is non-trivial. Moreover, \Approach only uses images for unlabeled data, avoiding the significant setup costs associated with collecting multi-sensor data.

\myheading{2D foundation models for depth estimation and segmentation.} Segmentation models like Segment Anything \cite{SegmentAnything} have introduced class-agnostic segmentation, demonstrating robustness across diverse datasets. Following this, models such as X-Decoder \cite{X-Decoder}, SEEM \cite{SEEM}, and OpenSEED \cite{OpenSEED} have emerged as general-purpose panoptic segmentation tools, offering zero-shot application on autonomous datasets. For depth estimation, models like Metric3Dv2 \cite{Metric3D} and UniDepth \cite{piccinelli2024unidepth} excel in robust metric depth estimation by leveraging large-scale datasets. In our work, we utilize these 2D VFMs to generate 3D clues, addressing the challenge of limited labeled data.

\begin{figure*}[t]
  \centering
  \includegraphics[width=.96\textwidth]{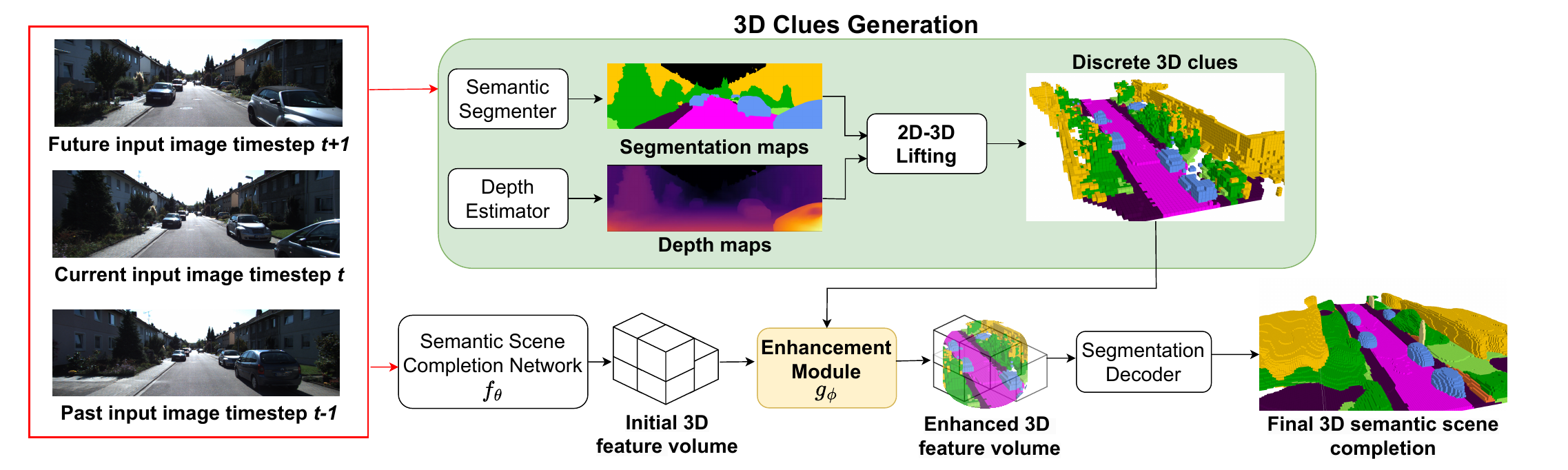}
  \caption{The overall architecture of the proposed VFG-SSC network. Our approach leverages 3D cues from 2D foundation models to enhance the inferred 3D feature volume and generate the final 3D semantic occupancy grid. The model is trained using all frames (solid red box) to produce pseudo-labels for the unlabeled data.}
  \label{fig:overviewSSC}
\end{figure*}

\section{Semi-Supervised Semantic Scene Completion}
\label{sec:approach}

\myheading{Problem Setting:}
Given a sequence $\mathbf{Q}_t = \{I_{t-k}, I_{t-k+1}\ldots, I_{t-1}, I_t \}$ of $k$ consecutive frames, the SSC model $f_\theta$ generates a semantic occupancy grid which is defined in the coordinate system of the ego vehicle at the timestamp $t$. Each voxel of the grid is categorized as either empty or occupied by a specific semantic class. The grid can be obtained as follows: 
$\mathbf{\hat{Y}}_t = f_\theta(\mathbf{Q_t})$ where $\mathbf{\hat{Y}}_t \in \mathbb{R}^{H \times W \times Z \times (C+1)}$. $H$, $W$, and $Z$ denote the voxel grid's height, width,
and depth, and $C$ is the number of the semantic classes. In a semi-supervised setting (\Approach), the dataset contains two non-overlapping subsets:
\begin{itemize}
    \item Labeled data $\mathcal{D}^L=\{(\mathbf{Q}_t, \mathbf{Y}_t)\}_{t=1}^L$ where each image $I_t$ has a corresponding 3D occupancy ground-truth $\mathbf{Y}_t$.
    \item Unlabeled data $\mathcal{D}^U = \{\mathbf{Q}_t\}_{t=1}^U$, which only contains images with \textbf{no LiDAR} and the amount of the unlabeled data is significantly larger than labeled data, i.e, $U \gg L$.
\end{itemize}

\begin{table}
    \centering
    \setlength{\tabcolsep}{1.2pt}

    \begin{tabular}{lccc}
    \toprule
    \textbf{Methods} & \textbf{Precision $\uparrow$} & \textbf{Recall $\uparrow$} & \textbf{mIoU $\uparrow$} \\
    \midrule
    Sup-only & 47.53 & 47.09 & 9.16\\
    Self-Training (ST) & 42.83 & \textbf{59.35} & 9.48\\
    3D clues as pseudo-labels & \textbf{48.60} & 26.19 & 9.27\\
    Our approach & 48.01 & 58.36 & \textbf{11.73} \\

    \bottomrule
    \end{tabular}
    \caption{Comparison of \Approach~on 5\% labeled SemanticKITTI validation data. The Sup-only baseline excludes pseudo-labels from unlabeled data during training.}
    \label{tab:precision-recall}
    
\end{table}

\myheading{Our Customized Self-Training for \Approach:}
\label{self-training}
A common strategy for tackling the semi-supervised problem is Self-Training. This involves first training a supervised model $f_\theta$ on labeled data $\mathcal{D}^L$ and then generating pseudo-labels for the unlabeled dataset $\mathcal{D}^U$. After that, the model $f_\theta$ is retrained using both the labeled and pseudo-labeled data.
Based on this Self-Training approach, our key contribution is to enhance the quality of the pseudo-label by incorporating 3D priors extracted from 2D vision foundation models. We now outline our three-step process in detail.



%


\begin{figure}[tb]
  \centering
  \includegraphics[width=\columnwidth]{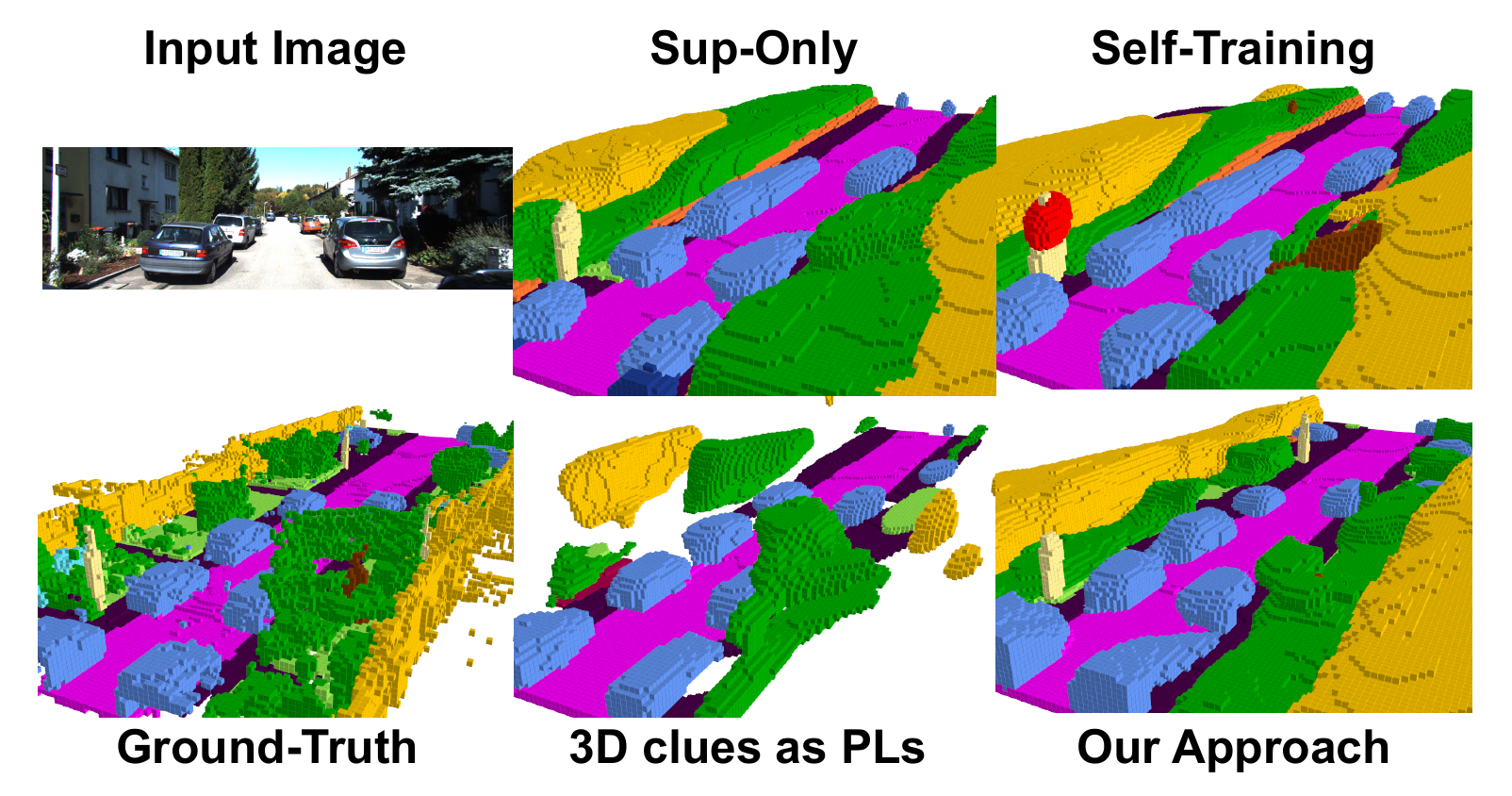}
  \caption{Qualitative of \Approach~approaches. With Sup-only and Self-Training, the scene layout is reasonably reconstructed, however, dynamic object prediction is incorrect due to many false positives. With 3D clues as pseudo-labels, objects are correctly predicted but occluded regions do not have a prediction (assigned as empty). Our method obtains reliable predictions (high precision) and reasonable reconstruction in occluded space (high recall).}
  \label{fig:fusing_fig}
\end{figure}

\begin{itemize}
    \item \textbf{Step 1 - Training on labeled data}: We first extract efficient 3D cues from 2D images using 2D foundation models. These cues are fused with 3D features derived from a 3D SSC network $f_\theta$ through our proposed enhancement module $g_\phi$. The networks $f_\theta$ and $g_\phi$ are jointly trained on labeled data $\mathcal{D}^L$, enabling the proposed enhancement module to learn the efficient fusion between 3D clues and 3D features with the supervision of labeled data. 
    
    \item \textbf{Step 2 - Generating pseudo-label for unlabeled data}: The trained $f_\theta$ and $g_\phi$ from Step 1 are used to predict unlabeled data $\mathcal{D}^U$, resulting in a pseudo-labeled dataset $\hat{\mathcal{D}}^L=\{(\mathbf{I}_t, \hat{\mathbf{Y}}_t )\}_{t=1}^U$. 
    
    \item \textbf{Step 3 - Retraining with labeled and pseudo-labeled data}: Finally, we retrain the original SSC model $f_\theta$ using the combined dataset $\mathcal{D}^L \cup \hat{\mathcal{D}}^L$ with supervised segmentation losses. Only $f_{\theta}$ is utilized during test time; 3D cues and $g_{\phi}$ are not used at this stage.
\end{itemize}

The next part presents our contributions in Steps 1 and 2. 

\section{Our VFG-SSC Network}

To highlight the importance of improving pseudo-label quality in Self-Training, we conducted a quick test by training $f_{\theta}$ with various setups, as shown in \cref{tab:precision-recall}. As can be seen, using additional pseudo-labeled data from Self-Training (second row) boosts recall but lowers precision compared to the model trained on 5\% labeled data only (first row). 
To overcome this limitation, we incorporate 3D cues from 2D VFMs as pseudo labels for the unlabeled data. These 3D cues offer higher precision but lower recall, as they provide accurate yet sparse estimates of depth and semantics. By combining these sources, we enhance both precision and recall, leading to an overall improvement in mIoU. This approach motivates our strategy to align 3D cues with Self-Training by integrating them as inputs during pseudo-label generation. We achieve this enhancement through an attention-based mechanism that effectively fuses the two feature sets.

Our VFG-SSC network, as shown in ~\cref{fig:overviewSSC}, includes a unique 3D clue generation module, a semantic scene completion network $f_\theta$, and an enhancement module $g_\phi$.

\subsection{3D Clues Generation} \label{sec:generation}

\begin{figure}[t]
  \centering
  \includegraphics[width=\columnwidth]{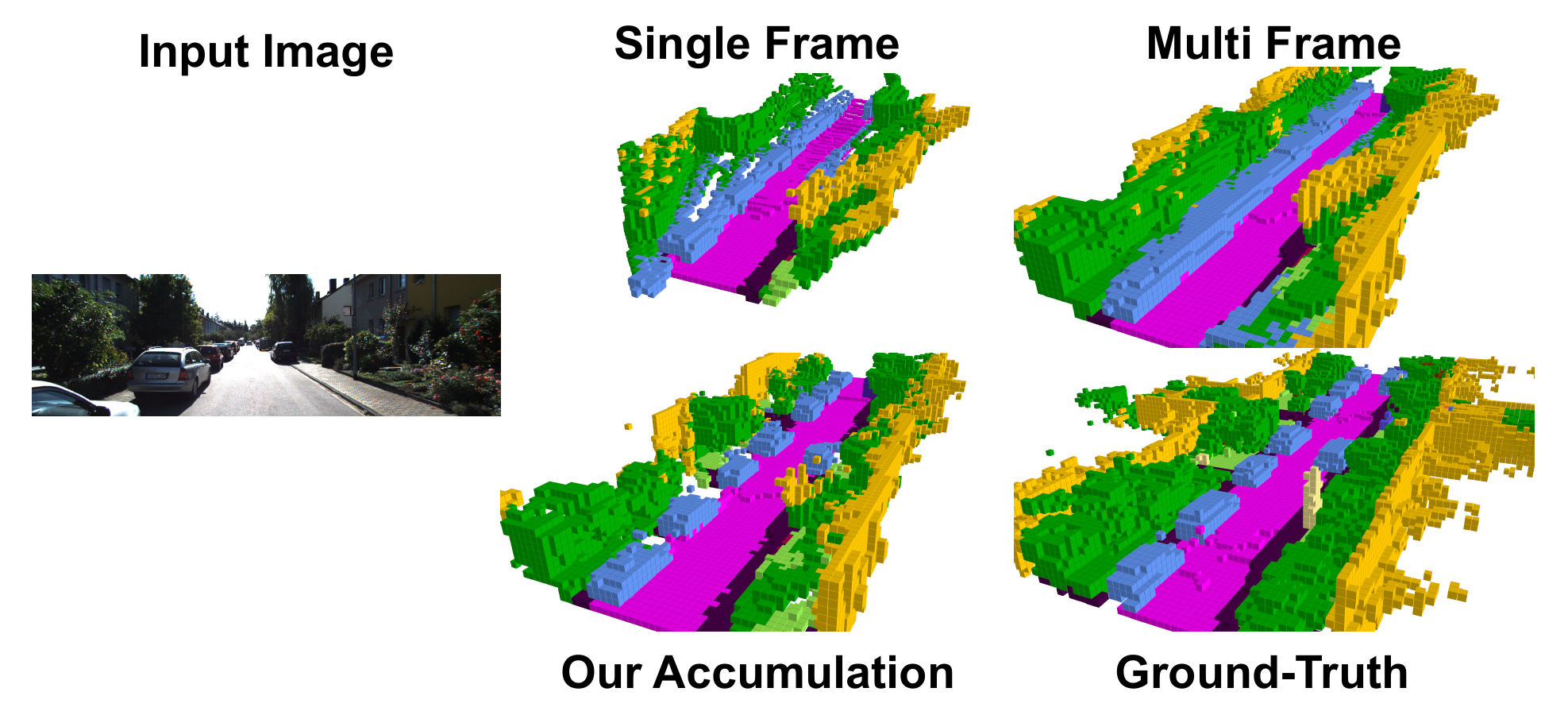}
  \caption{3D clues visual comparisons on different settings: using only the current image and accumulated 3D clues from temporal frames with and without filtering.}
  \label{fig:filtering_choice}
\end{figure}

As mentioned in the previous section, we exploit 2D VFMs of depth estimation and semantic segmentation to create 3D clues. A 3D clue is defined as a discrete set of voxels $\mathbf{Y}_{\text{clues}} \in \mathbb{R}^{H \times W \times Z \times (C+1)}$. A straightforward approach is to extract depth and semantic information from the current frame, and then use the depth data to project the segmentation into a 3D voxel representation. However, as shown in ~\cref{fig:filtering_choice}, 3D clues generated from a single frame often contain numerous irrelevant artifacts that do not align with the actual scene. To address this issue, we propose generating clues using temporal image sequences. Particularly,
given a set of $N$ input images 
of a scene, we extract depth maps 
and semantic segmentation map from VFMs. 
For each pixel $u = [u_x, u_y]$ and its corresponding depth $d$, we un-project the semantic map $u$ to 3D point $x$ in world coordinate by computing $x = E^{-1} \times K^{-1} \times [u_x, u_y, d]^T$, 
where $K$ and $E$ are the camera intrinsic and extrinsic matrices respectively. 
This results in a set of semantic point clouds $X \in \mathbb{R}^{N\times 4}$ (3D coordinates and semantic class). We convert these point clouds to voxels to obtain $\mathbf{Y}_{\text{clues}}$.

As shown in \cref{fig:filtering_choice}, accumulating 3D clues from temporal frames (Multi Frame) results in a more accurate scene representation compared to using a single frame. However, dynamic objects like vehicles become indistinguishable. To address this, we introduce a regularization pipeline that leverages semantic and geometric constraints. We begin by segmenting the semantic point cloud produced in the previous step by class and applying heuristic filters: a radius filter to remove outliers based on the density of nearby points, and a statistical filter to eliminate points that deviate significantly from the average distance. The results in \cref{tab:Components} show that this pipeline is crucial for effectively utilizing VFMs. A pseudo-code implementation is provided in the \textit{Supplementary Material}.

\subsection{3D Clue-Guided Enhancement Module}
\label{sec:fusion}

This module, $g_\phi$, fuses the 3D clues from the previous module with the 3D features generated by the scene completion network $f_\theta$ to create an enhanced 3D feature volume. A naive approach is to concatenate these 3D clues with the features from $f_\theta$ to obtain a refined feature. However, our visualized results in \cref{fig:fusing_fig} reveal that these features contain complementary information that can be better utilized. The 3D clues can reliably estimate the semantics and geometry of visible surfaces, but due to the limited camera FoV, many voxels remain without sufficient information. Concatenation alone may introduce ambiguities in these regions. An alternative approach involves cross-attention, which associates each feature in $f_{\theta}$ with all 3D clues, but this can lead to a significant increase in computational cost. To balance effectiveness and efficiency, we propose using a local attention mechanism to merge the two features, achieving both global context aggregation and linear computational complexity.

Our enhancement module uses a set of self-attention blocks to align the two features implicitly. Specifically, given 3D clues $\mathbf{Y}_{\text{clues}}$ and 3D features $\mathbf{Y}_{\text{feats}}$, we project the discretized 3D clues into continuous embeddings. We then concatenate the aligned features with the original 3D features and apply self-attention for enhancement. Note that we adopt Dilated Neighborhood Attention \cite{hassani2023neighborhood}, a recently proposed method with linear time and space complexity, to reduce the computational cost of self-attention.

\begin{table*}[tb]
\setlength{\tabcolsep}{6pt}
\centering
\begin{tabular}{l l c c c c c c}
\toprule
\multirow{2}{*}{\textbf{Backbone}} & \multirow{2}{*}{\textbf{Method}} & \multicolumn{2}{c}{\textbf{1\%}} & \multicolumn{2}{c}{\textbf{5\%}} & \multicolumn{2}{c}{\textbf{10\%} } \\
&                          & mIoU$\uparrow$ & IoU$\uparrow$    & mIoU$\uparrow$  & IoU$\uparrow$    & mIoU$\uparrow$  & IoU$\uparrow$    \\

\midrule
    \rowcolor{gray!70} SceneRF  & Self-supervised & \multicolumn{6}{c}{N/A mIoU, 13.84 IoU} \\
    \rowcolor{gray!70} SelfOcc & Self-supervised  & \multicolumn{6}{c}{5.02 mIoU, 21.97 IoU} \\
\midrule
SelfOcc & Finetuned  & 5.55 & 29.57 & 7.5 & 32.43 & 8.3 & 33.66 \\
\midrule
     \rowcolor{gray!70}  \multirow{6}{*}{{OccFormer}}  &  Fully-Supervised   & \multicolumn{6}{c}{14.08 mIoU, 36.04 IoU}    \\
                        & Supervised-only       & 6.59 & 26.31 & 9.16  & 32.73 & 10.45 & 34.25 \\
                        & Mean-Teacher          & 6.76 & 26.54 & 9.41  & 31.67 & 10.53 & 33.28 \\
                        & Self-Training         & 6.75 & 26.61 & 9.48  & 33.12 & 10.49 & 33.94 \\
                         & VFG-SSC (Ours)  & \textbf{9.40} & \textbf{30.40} & \textbf{11.73} & \textbf{35.76} & \textbf{12.38} & \textbf{35.25} \\
\midrule
     \rowcolor{gray!70} \multirow{5}{*}{{VoxFormer}} &  Fully-Supervised         & \multicolumn{6}{c}{14.40 mIoU, 43.50 IoU}      \\
                        & Supervised-only   & 6.35 & 34.04 & 9.56  & 39.68 & 10.52 & 40.28 \\
                        & Mean-Teacher      & 7.20  & \textbf{38.97} & 9.58  & 37.70  & 10.56 & 38.66 \\
                        & Self-Training    & 7.60  & 35.40 & 10.51 & 39.83 & 11.06 & 41.05 \\
                        & VFG-SSC (Ours)  & \textbf{9.32} & 36.78 & \textbf{11.15} & \textbf{39.96} & \textbf{12.19} & \textbf{41.57} \\
\bottomrule
\end{tabular}
\caption{Results on the val set of SemanticKITTI. \colorbox{gray!70}{Faded rows} indicate reference-only results, not for direct comparison.}
\label{tab:semanticKITTIquantitative}
\vspace{-10pt}
\end{table*}

\section{Experiments}
\label{sec:experiments}

\myheading{Datasets and metrics:} We utilize two 3D semantic occupancy datasets, SemanticKITTI \cite{SemanticKITTI} and NYUv2~\cite{silberman2012indoor}, which represent outdoor and indoor environments, respectively, to establish new semi-supervised benchmarks. For SemanticKITTI, we sample 40, 198, and 383 frames (corresponding to 1\%, 5\%, and 10\% of the training set), consistent with existing setups \cite{OpenOccupancy, SemanticKITTI}. For NYUv2, we uniformly sample 40 and 80 frames (representing 5\% and 10\% of the training set). We evaluate performance using mIoU for semantic class and IoU for binary occupancy predictions.

\myheading{Implementation details:} 
For SemanticKITTI, we use OccFormer \cite{OccFormer} and VoxFormer \cite{VoxFormer} as our primary SSC networks ($f_\theta$). 
VoxFormer employs a two-stage training pipeline: we pre-train the query proposal with 1/5/10\% labeled data in stage 1, then apply VFG-SSC in stage 2. For OccFormer, which uses additional depth supervision from LiDAR, we supplement the missing LiDAR data in unlabeled sets with pseudo-LiDAR generated from a depth model \cite{piccinelli2024unidepth}. We maintain consistent settings across both models for fair comparison.
For NYUv2, we use MonoScene \cite{MonoScene} and OccDepth \cite{OccDepth} as the SSC networks. 

UniDepth \cite{piccinelli2024unidepth} is used for depth estimation, and SEEM \cite{SEEM} for semantics, serving as our foundational models. We employ 2 layers of Dilated Neighborhood Attention with 4 heads, a kernel size of 7, and 4 dilation rates (1, 2, 4, 8). Additional details on losses for training each SSC network and further implementation specifics are provided in the Supplementary Material.

\begin{table*}
\small
\setlength\tabcolsep{3pt} 
\centering
\begin{tabular}{lccccccccccccccccccccc}
\toprule
\textbf{Method} &
  \textbf{mIoU$\uparrow$ } &
  \textbf{IoU$\uparrow$ } &
  \rotatebox[origin=c]{90}{\textbf{road}} &
  \rotatebox[origin=c]{90}{\textbf{sidewalk}} &
  \rotatebox[origin=c]{90}{\textbf{parking}} &
  \rotatebox[origin=c]{90}{\textbf{other-grnd}} &
  \rotatebox[origin=c]{90}{\textbf{building}} &
  \rotatebox[origin=c]{90}{\textbf{car}} &
  \rotatebox[origin=c]{90}{\textbf{truck}} &
  \rotatebox[origin=c]{90}{\textbf{bicycle}} &
  \rotatebox[origin=c]{90}{\textbf{motorcycle}} &
  \rotatebox[origin=c]{90}{\textbf{other-vehi}} &
  \rotatebox[origin=c]{90}{\textbf{vegetation}} &
  \rotatebox[origin=c]{90}{\textbf{trunk}} &
  \rotatebox[origin=c]{90}{\textbf{terrain}} &
  \rotatebox[origin=c]{90}{\textbf{person}} &
  \rotatebox[origin=c]{90}{\textbf{bicyclist}} &
  \rotatebox[origin=c]{90}{\textbf{motorcyclist}} &
  \rotatebox[origin=c]{90}{\textbf{fence}} &
  \rotatebox[origin=c]{90}{\textbf{pole}} &
  \rotatebox[origin=c]{90}{\textbf{traffic-sign}} \\ 
\midrule
\rowcolor{gray!70} MonoScene &
  11.1 &
  34.2 &
  54.7 &
  27.1 &
  24.8 &
  5.7 &
  14.4 &
  18.8 &
  3.3 &
  0.5 &
  0.7 &
  4.4 &
  14.9 &
  2.4 &
  19.5 &
  1.0 &
  1.4 &
  0.4 &
  11.1 &
  3.3 &
  2.1 \\
  \rowcolor{gray!70} OccFormer  &
  13.2 &
  33.7 &
  57.3 &
  31.1 &
  29.7 &
  13.7 &
  16.8 &
  21.5 &
  4.5 &
  2.4 &
  2.2 &
  3.2 &
  15.5 &
  4.3 &
  22.5 &
  2.7 &
  2.3 &
  0.0 &
  12.5 &
  4.1 &
  4.2 \\
\rowcolor{gray!70} VoxFormer & 14.3 &
  42.0 &
  56.5 &
  30.0 &
  25.8 &
  12.3 &
  23.7 &
  23.4 &
  5.1 &
  3.4 &
  1.6 &
  2.7 &
  24.5 &
  8.6 &
  24.4 &
  1.6 &
  1.3 &
  0.0 &
  15.4 &
  6.3 &
  6.0\\
VFG-SSC (Occ) &
  11.1 &
  30.5 &
  55.2 &
  26.7 &
  25.6 &
  0.0 &
  14.8 &
  20.5 &
  1.6 &
  2.3 &
  2.2 &
  5.8 &
  13.1 &
  3.6 &
  20.7 &
  3.4 &
  0.0 &
  0.0 &
  10.5 &
  3.4 &
  2.1 \\
VFG-SSC (Vox) &
  11.8 &
  40.6 &
  55.5 &
  23.8 &
  16.3 &
  0.0 &
  20.9 &
  21.5 &
  2.2 &
  1.4 &
  1.8 &
  3.5 &
  23.7 &
  8.2 &
  20.0 &
  1.5 &
  0.0 &
  0.0 &
  12.5 &
  5.8 &
  4.9 \\
\bottomrule
\end{tabular}%
\caption{Quantitative comparisons between our method and other fully-supervised methods on the hidden testing set of SemanticKITTI~\cite{SemanticKITTI} dataset. \colorbox{gray!70}{Faded rows} denote results for reference purposes only, not direct comparison.}
\label{tab:SemanticKITTItestset}

\end{table*}

\subsection{Comparison with Related Methods}

As the first work to address the 3D semantic scene completion task in a semi-supervised manner, we adapt two semi-supervised learning approaches as baselines: Mean Teacher \cite{MeanTeacher}, which leverages consistency regularization, and Self-Training \cite{Self-Training} with entropy minimization.
In the Semi-SSC setting, we train our model using only 1\%, 5\%, and 10\% of the labeled training data and compare its performance against other methods trained with the same data percentages. For reference, we also include results from fully supervised training using 100\% labeled data, as well as two self-supervised methods, SceneRF~\cite{cao2023scenerf} and SelfOcc~\cite{huang2023selfocc}, which rely entirely on unlabeled data. We also evaluate a version of SelfOcc fine-tuned on the same labeled data percentages with an added segmentation head.

\begin{table}[tb]
\small
\centering
\setlength\tabcolsep{3pt} 
\begin{tabular}{l l c c c c }
\toprule
\multirow{2}{*}{\textbf{SSC Network}} & \multirow{2}{*}{\textbf{Method}} & \multicolumn{2}{c}{\textbf{5\%}} & \multicolumn{2}{c}{\textbf{10\%} } \\
                    &                          & \textbf{mIoU$\uparrow$}  & \textbf{IoU$\uparrow$}    & \textbf{mIoU$\uparrow$}  & \textbf{IoU$\uparrow$}    \\
\midrule
\rowcolor{gray!70} \multirow{5}{*}{{MonoScene}} &  Fully-Supervised         & \multicolumn{4}{c}{26.94 mIoU, 42.51 IoU}    \\
                    & Supervised-only                 & 13.36 & 29.17 & 17.17 & 34.68 \\
                    & Self-Training           & 14.00  & 30.64 & 17.32 & 35.20 \\
                    & VFG-SSC  & \textbf{18.23 } & \textbf{43.35} & \textbf{22.20 } & \textbf{44.37} \\
\midrule
\rowcolor{gray!70} \multirow{5}{*}{{OccDepth}} &  Fully-Supervised         & \multicolumn{4}{c}{29.03 mIoU, 44.17 IoU}      \\
                    & Supervised-only                 & 16.69  & 30.97 & 19.01 & 37.83 \\
                    & Self-Training    & 16.72 & 30.49 & 19.42 & 38.07 \\
                    & VFG-SSC & \textbf{22.54} & \textbf{43.61} & \textbf{23.92} & \textbf{44.85} \\
\bottomrule
\end{tabular}%
\caption{Quantiative results on the testing set of the  NYUv2~\cite{silberman2012indoor} dataset. \colorbox{gray!70}{Faded rows} denote results for reference purposes only, not direct comparison.}
\label{tab:NYUv2quantitative}

\end{table}

\begin{figure*}[t]
		\centering
		\newcolumntype{P}[1]{>{\centering\arraybackslash}m{#1}}
		\setlength{\tabcolsep}{0.001\textwidth}
		\renewcommand{\arraystretch}{0.8}
		\footnotesize
		\begin{tabular}{P{0.157\textwidth} P{0.157\textwidth} P{0.157\textwidth} P{0.157\textwidth} P{0.157\textwidth} P{0.157\textwidth}}		
			Input & Mean-Teacher & Self-Training  & SelfOcc  & Ours & Ground Truth
			\\[-0.1em]
			\includegraphics[width=\linewidth]{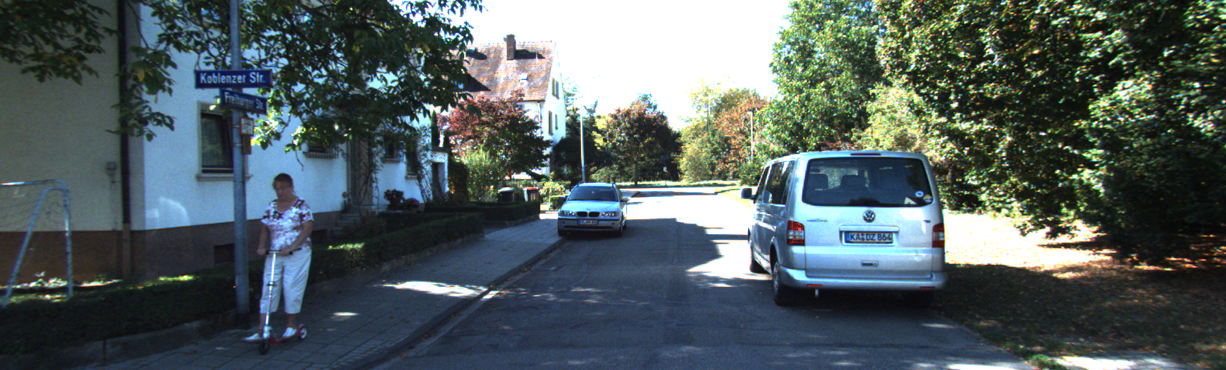} & 
			\includegraphics[width=\linewidth]{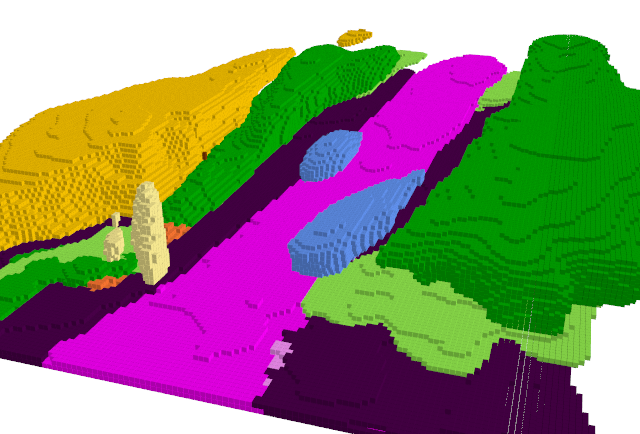} & 
			\includegraphics[width=\linewidth]{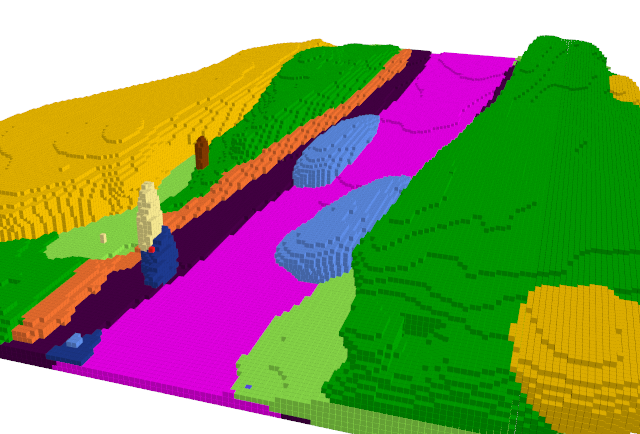} & 
			\includegraphics[width=\linewidth]{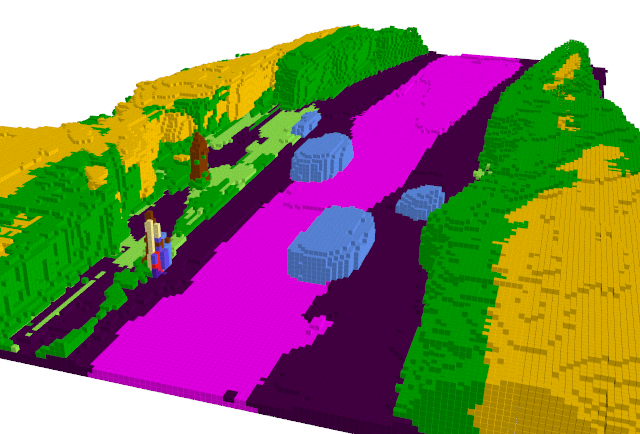} & 
			\includegraphics[width=\linewidth]{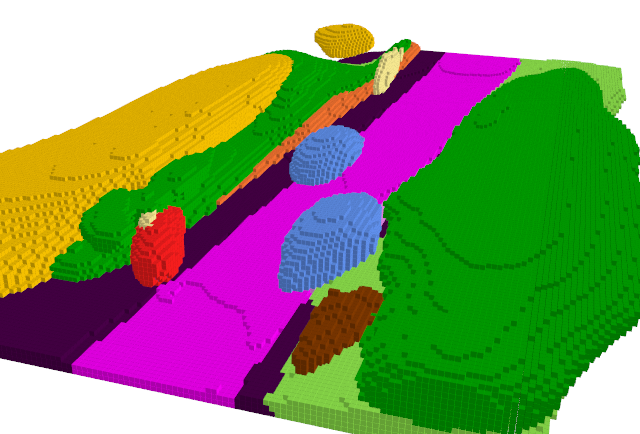} & 
			\includegraphics[width=\linewidth]{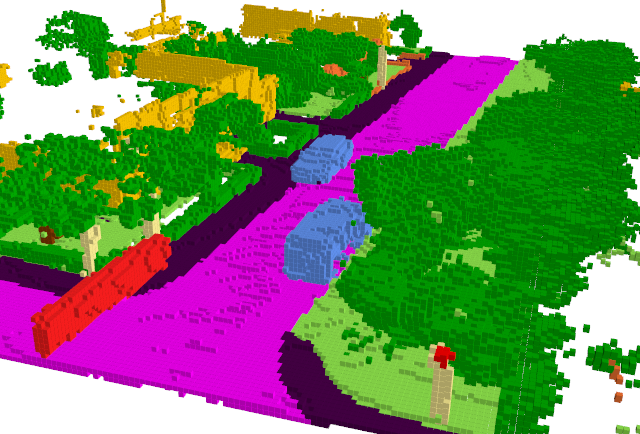} 
			\\[-0.6em]
			\includegraphics[width=\linewidth]{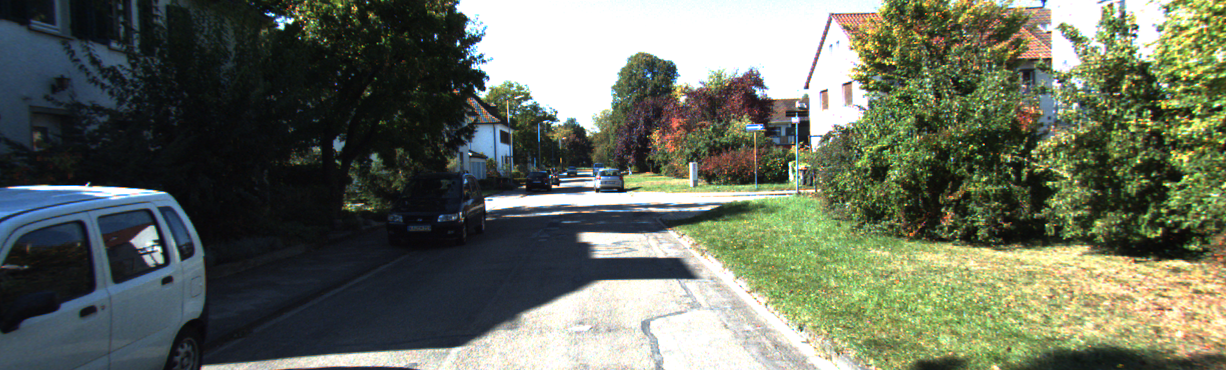} & 
			\includegraphics[width=\linewidth]{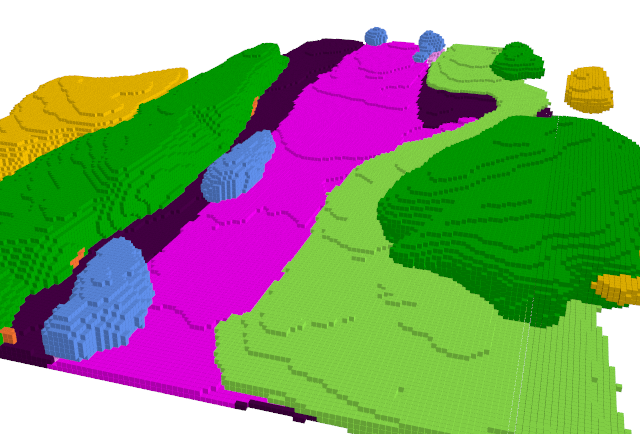} & 
			\includegraphics[width=\linewidth]{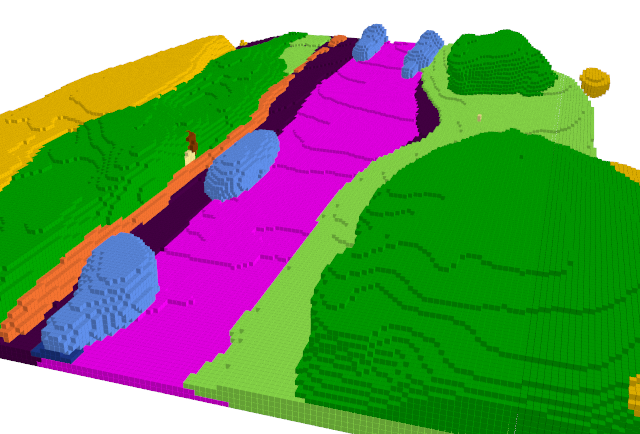} & 
			\includegraphics[width=\linewidth]{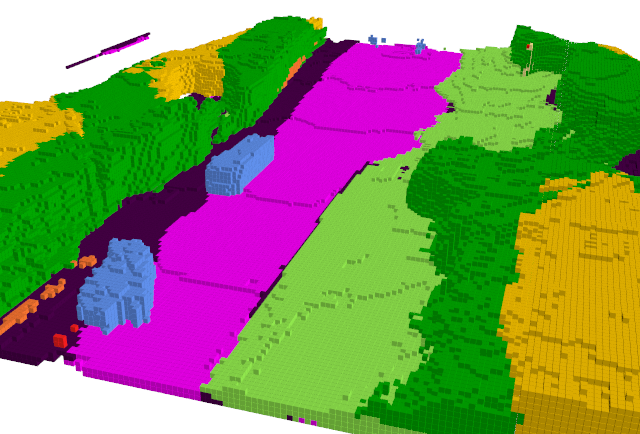} & 
			\includegraphics[width=\linewidth]{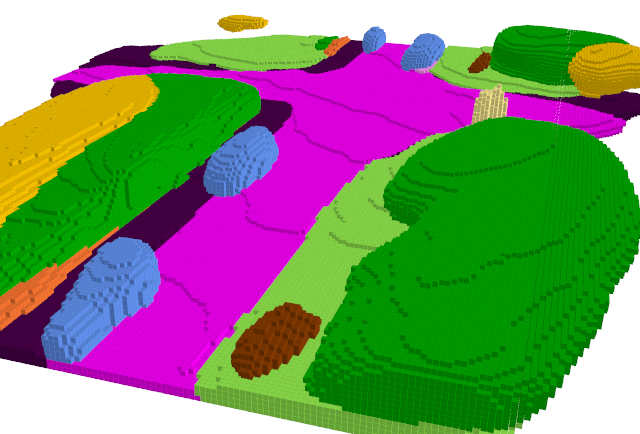} & 
			\includegraphics[width=\linewidth]{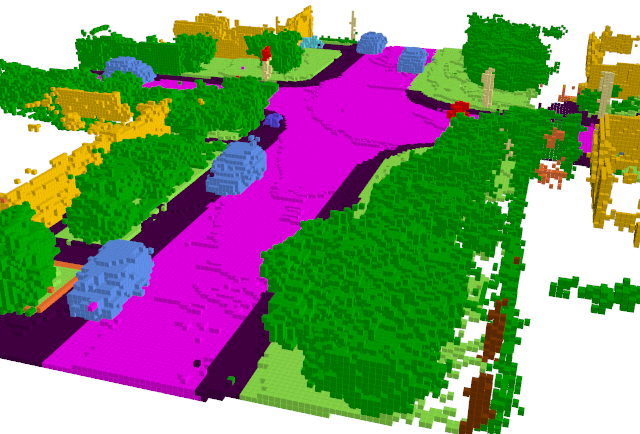} 
			\\[-0.6em]
			\includegraphics[width=\linewidth]{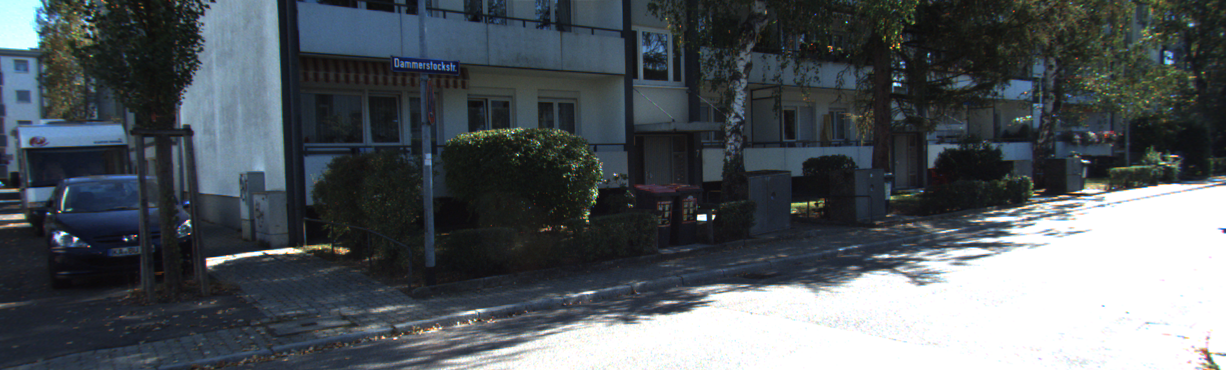} & 
			\includegraphics[width=\linewidth]{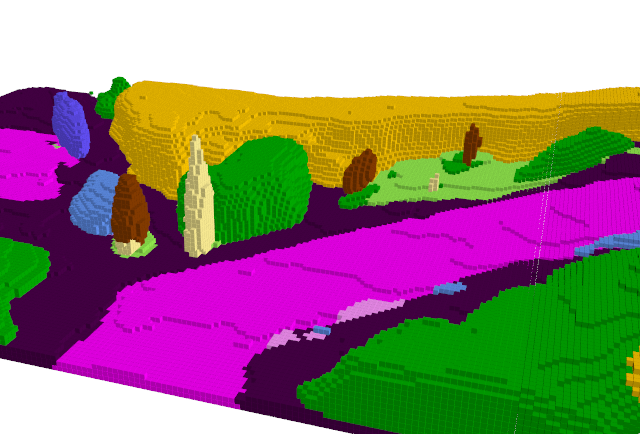} & 
			\includegraphics[width=\linewidth]{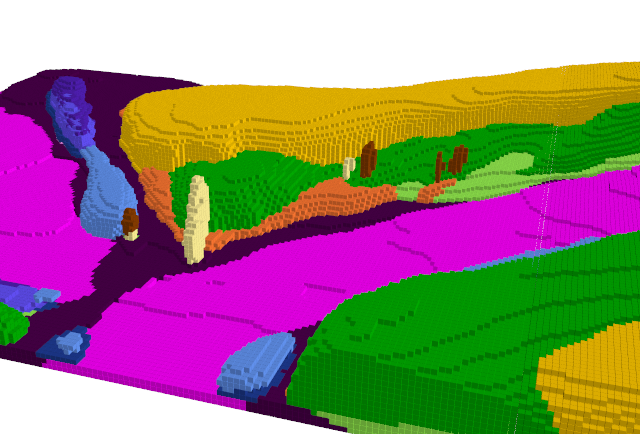} & 
			\includegraphics[width=\linewidth]{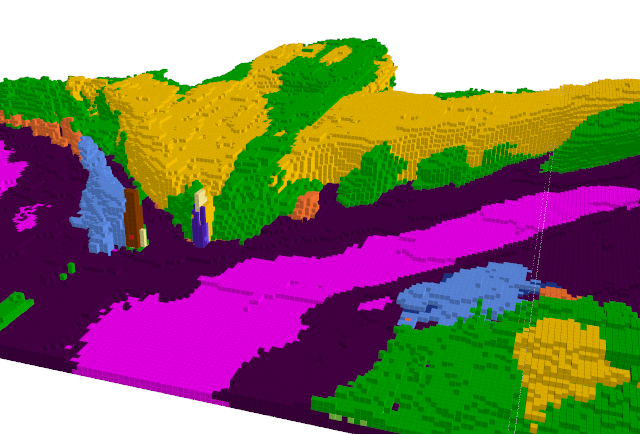} & 
			\includegraphics[width=\linewidth]{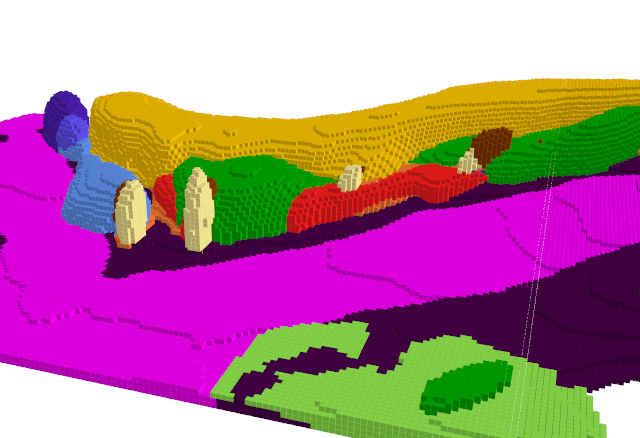} & 
			\includegraphics[width=\linewidth]{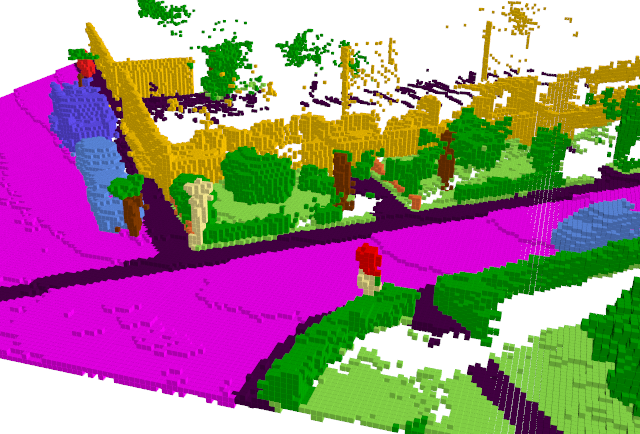} 
			\\		
                \multicolumn{6}{c}{
				\scriptsize
				\textcolor{bicycle}{$\blacksquare$}bicycle~
				\textcolor{car}{$\blacksquare$}car~
				\textcolor{motorcycle}{$\blacksquare$}motorcycle~
				\textcolor{truck}{$\blacksquare$}truck~
				\textcolor{other-vehicle}{$\blacksquare$}other vehicle~
				\textcolor{person}{$\blacksquare$}person~
				\textcolor{bicyclist}{$\blacksquare$}bicyclist~
				\textcolor{motorcyclist}{$\blacksquare$}motorcyclist~
				\textcolor{road}{$\blacksquare$}road~
				\textcolor{parking}{$\blacksquare$}parking~
                } \\
			\multicolumn{6}{c}{
				\scriptsize
				\textcolor{sidewalk}{$\blacksquare$}sidewalk~
				\textcolor{other-ground}{$\blacksquare$}other ground~
				\textcolor{building}{$\blacksquare$}building~
				\textcolor{fence}{$\blacksquare$}fence~
				\textcolor{vegetation}{$\blacksquare$}vegetation~
				\textcolor{trunk}{$\blacksquare$}trunk~
				\textcolor{terrain}{$\blacksquare$}terrain~
				\textcolor{pole}{$\blacksquare$}pole~
				\textcolor{traffic-sign}{$\blacksquare$}traffic sign			
			}
		\end{tabular}		
		  \caption{Qualitative results on the validation set of the SemanticKITTI~\cite{SemanticKITTI} dataset.}
            \label{fig:SemKITTI_results}
\end{figure*}

\begin{figure}[t]
		\centering
		\footnotesize
		\renewcommand{\arraystretch}{0.0}
		\setlength{\tabcolsep}{0.8pt}
		\newcolumntype{P}[1]{>{\centering\arraybackslash}m{#1}}
		\begin{tabular}{P{0.09\textwidth} P{0.2\columnwidth} P{0.2\columnwidth} P{0.2\columnwidth} P{0.205\columnwidth}}
			Input & Sup-only & Self Training & Ours & Ground Truth 
			\\
			\includegraphics[width=\linewidth]{images/nyu_qualitative/color/NYU0445_0000_color.jpg} & 
                \includegraphics[width=\linewidth]{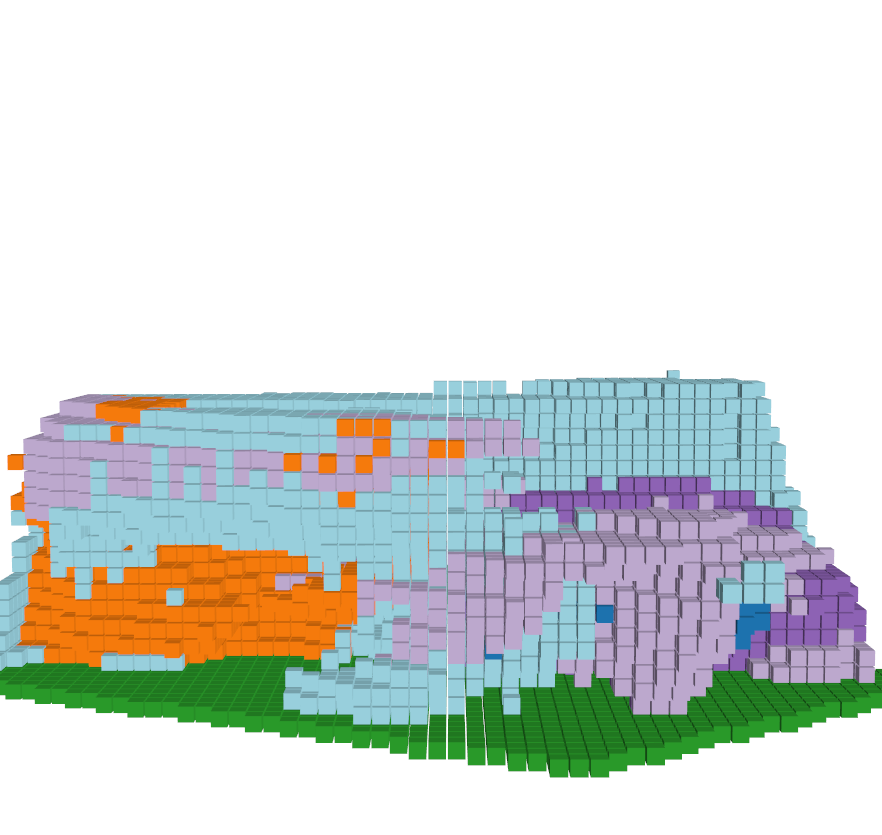} &
			\includegraphics[width=\linewidth]{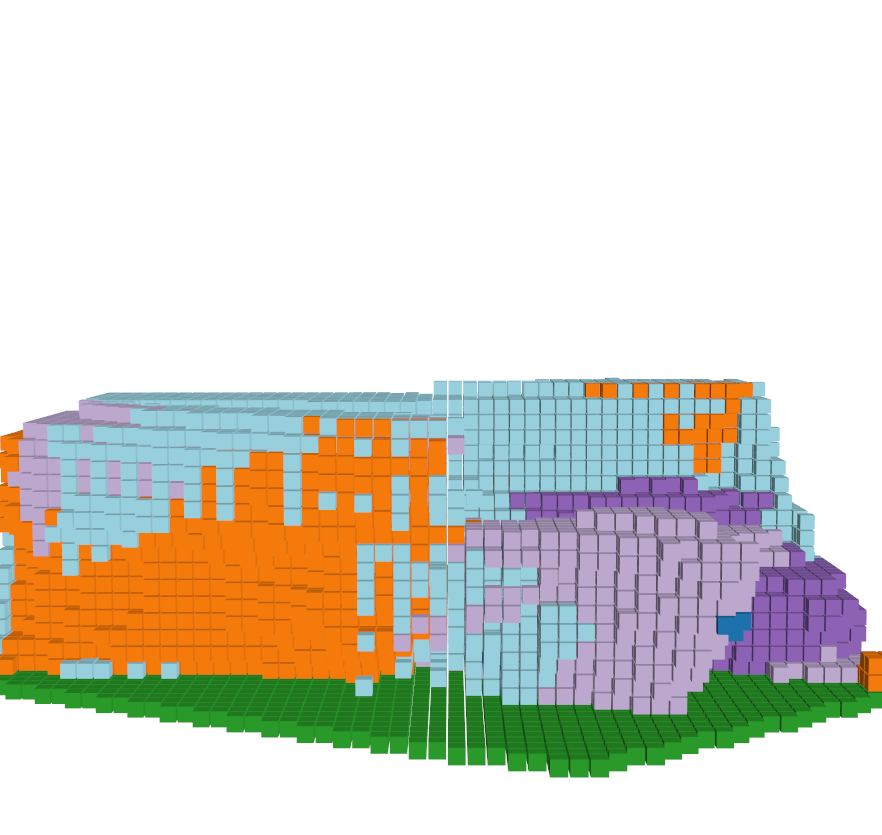} &
			\includegraphics[width=\linewidth]{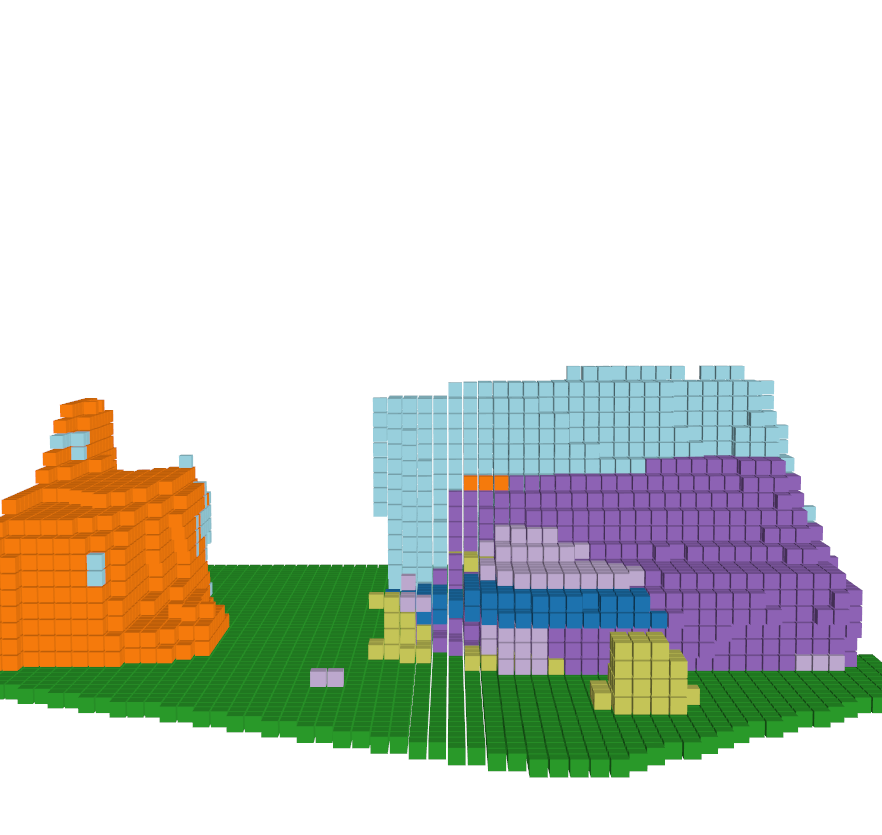} &
			\includegraphics[width=\linewidth]{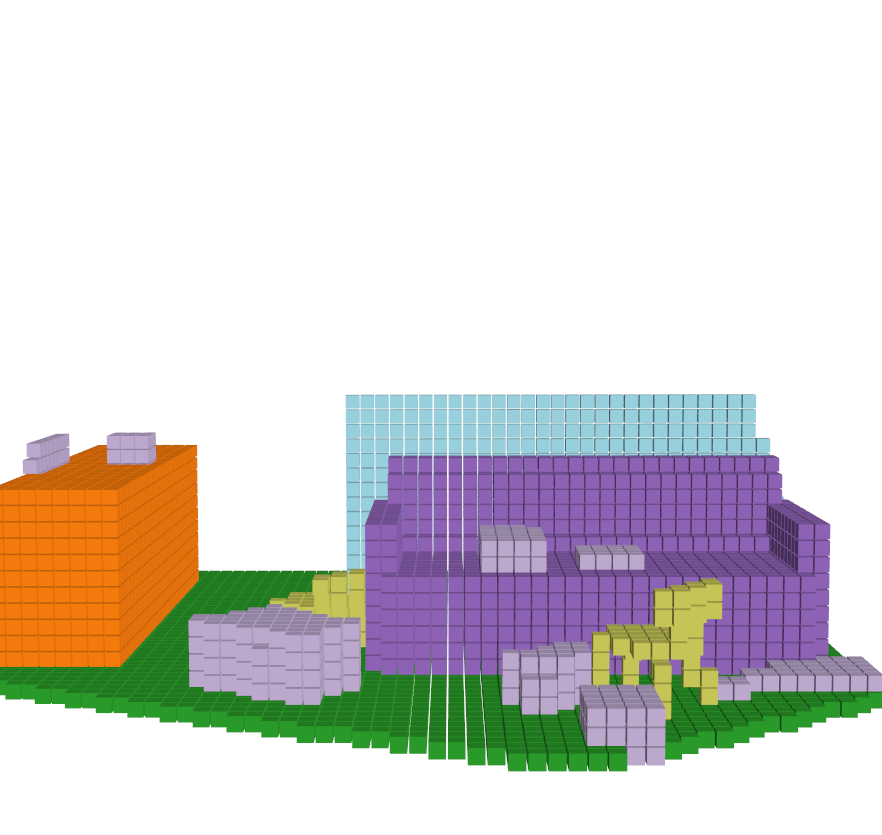} 
			\\[-0.4em]
			\includegraphics[width=\linewidth]{images/nyu_qualitative/color/NYU0618_0000_color.jpg} & \includegraphics[width=\linewidth]{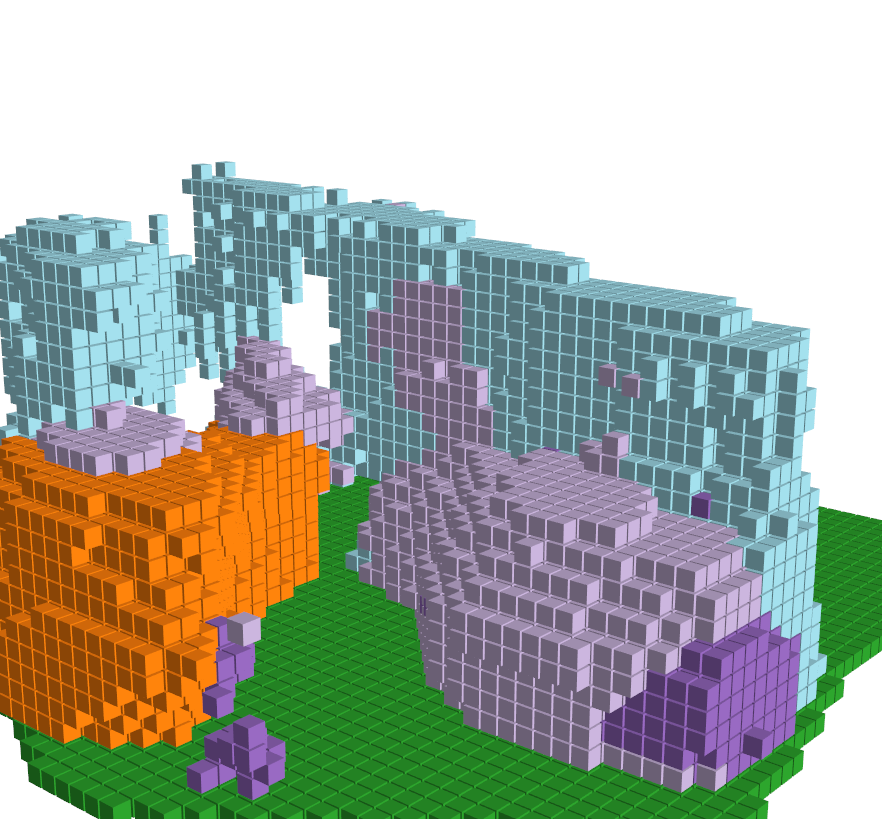} &
			\includegraphics[width=\linewidth]{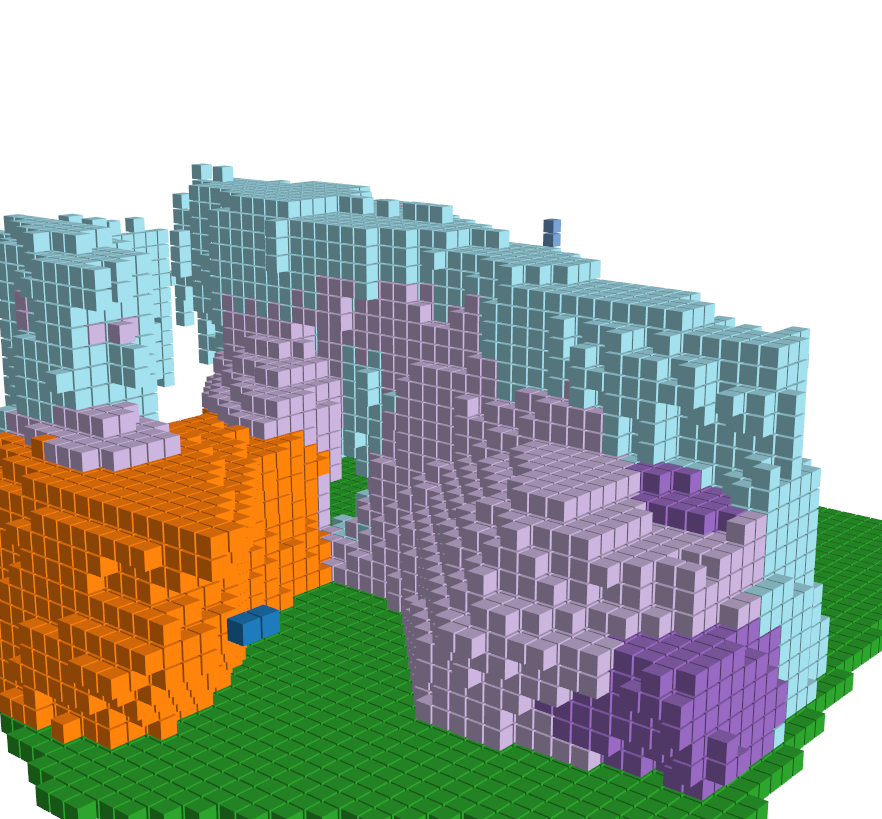} &
			\includegraphics[width=\linewidth]{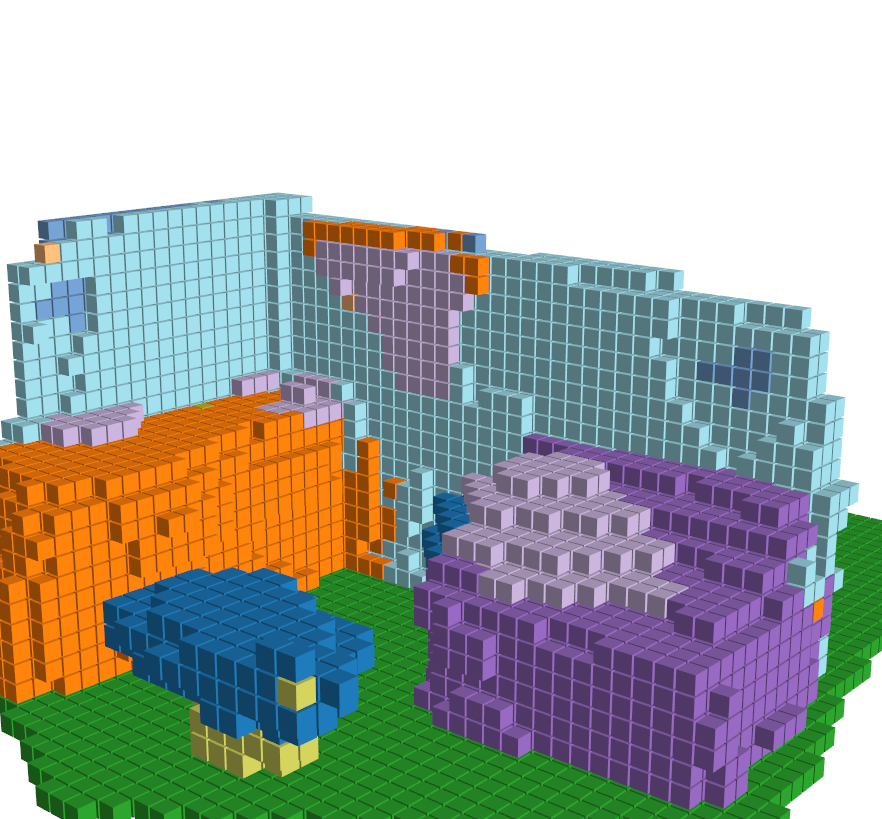} &
			\includegraphics[width=\linewidth]{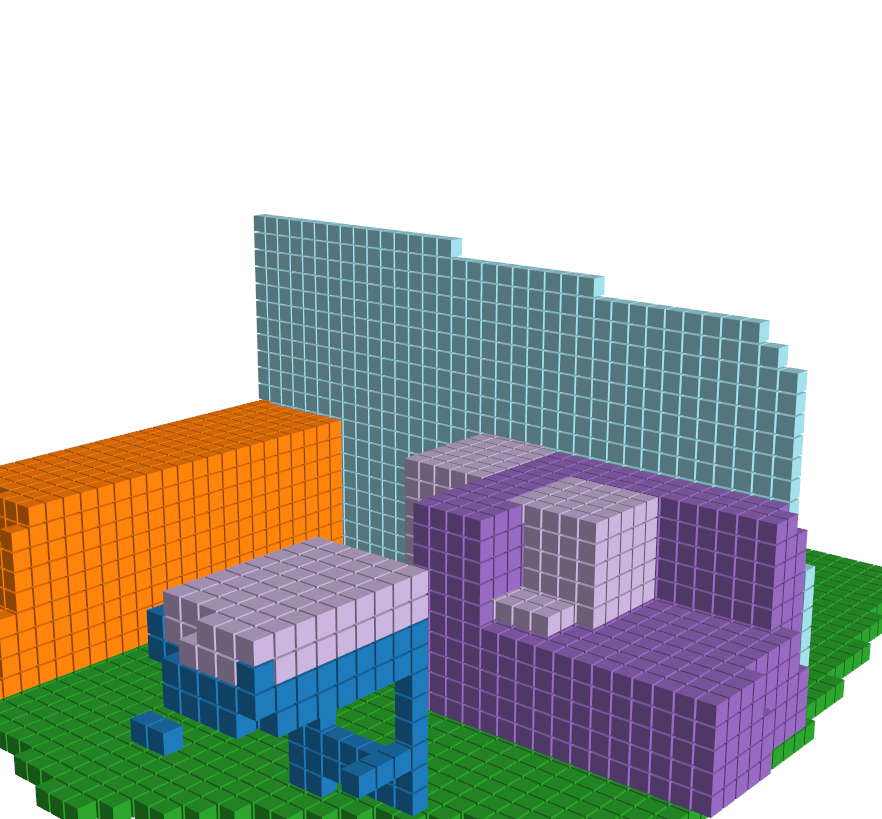} 
			\\[-0.3em] 
			\includegraphics[width=\linewidth]{images/nyu_qualitative/color/NYU1369_0000_color.jpg} & 
                \includegraphics[width=\linewidth]{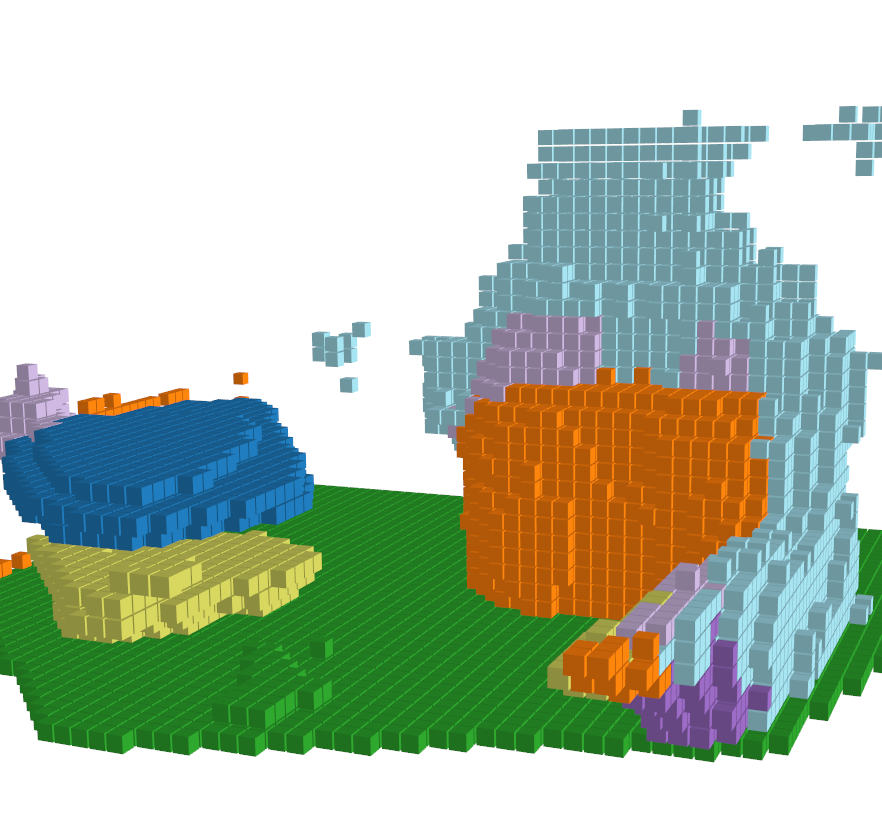} & 
			\includegraphics[width=\linewidth]{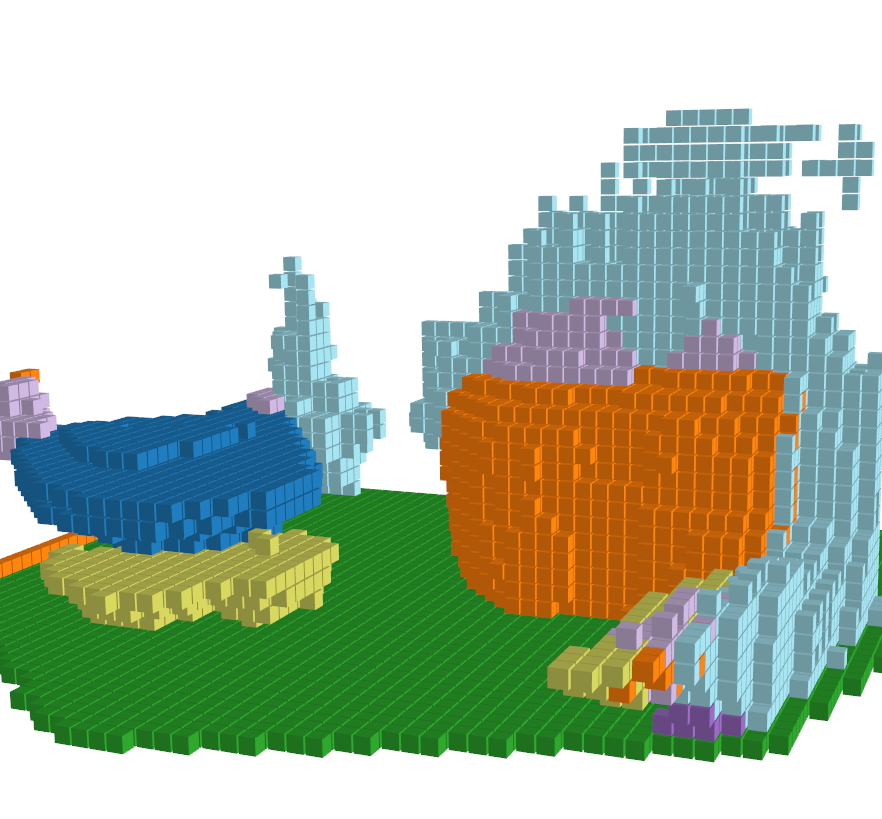} &
			\includegraphics[width=\linewidth]{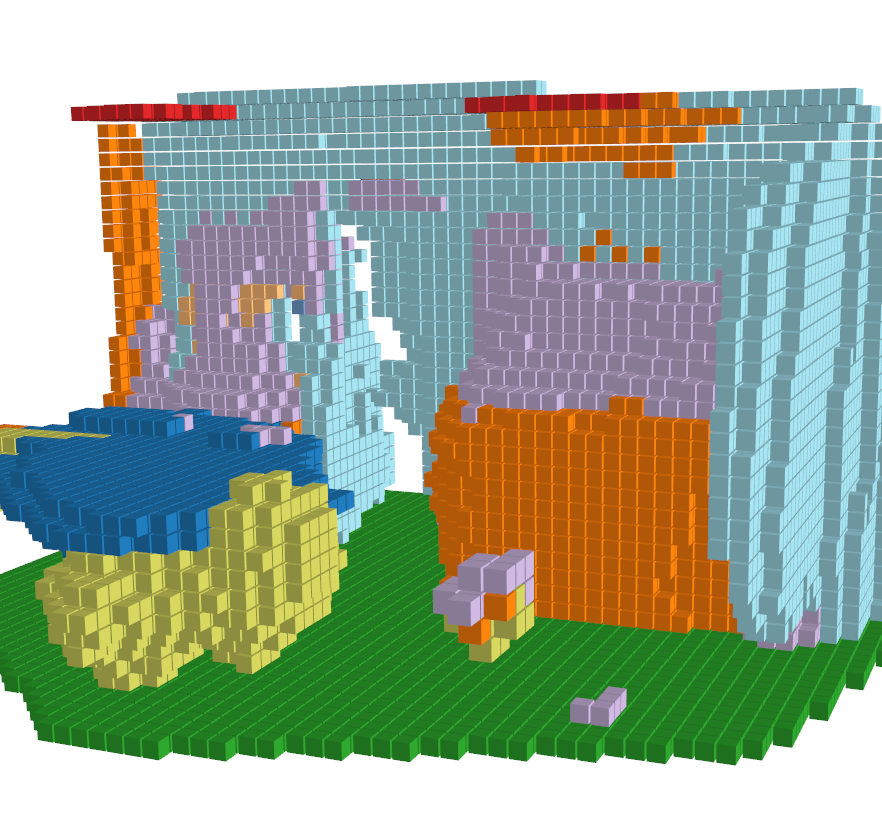} &
			\includegraphics[width=\linewidth]{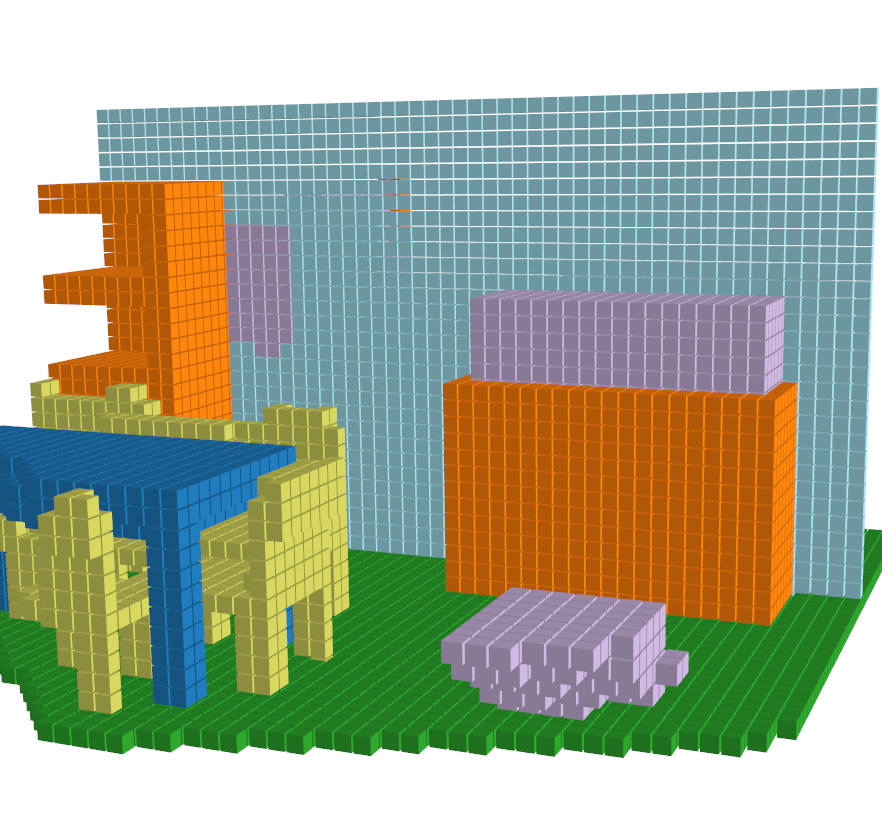} 
			\\
            \multicolumn{5}{c}{
				\scriptsize
				\textcolor{ceiling}{$\blacksquare$}ceiling~
				\textcolor{floor}{$\blacksquare$}floor~
				\textcolor{wall}{$\blacksquare$}wall~
				\textcolor{window}{$\blacksquare$}window~
				\textcolor{chair}{$\blacksquare$}chair~
				\textcolor{sofa}{$\blacksquare$}sofa~
                }
            \\
            \multicolumn{5}{c}{
                \scriptsize
				\textcolor{table}{$\blacksquare$}table~
				\textcolor{tvs}{$\blacksquare$}tvs~
				\textcolor{furniture}{$\blacksquare$}furniture~
				\textcolor{objects}{$\blacksquare$}objects
			}
		\end{tabular}
  \caption{Qualitative results on the testing set of the NYUv2~\cite{silberman2012indoor} dataset.}
  \label{fig:NYU_results}
\end{figure}

\myheading{Quantitative results:} For SemanticKITTI, we present IoU and mIoU in ~\cref{tab:semanticKITTIquantitative}. Additionally, we validate our results by reporting VFG-SSC performance with a 10\% labeled data setting on the SemanticKITTI hidden test set in Table ~\cref{tab:SemanticKITTItestset}. Our VFG-SSC significantly outperforms baseline methods. Remarkably, our approach achieves comparable results to fully supervised counterparts, with just a 15\% performance gap using 10\% labeled data. It is worth noting that on the 1\% setting, Mean Teacher obtains a better IoU score on VoxFormer. 
Therefore, a robust \Approach~method should achieve excellent performance in both geometric and semantic completion. 
Moreover, on the SemanticKITTI hidden test set, our method compares favorably with some fully supervised methods like MonoScene, despite utilizing only 10\% labeled occupancy annotation. This emphasizes the effectiveness of our 3D clues and enhancement module, which is effective in generating high-quality pseudo-labels for training any SSC backbones. For NYUv2, we report the results in ~\cref{tab:NYUv2quantitative}. We consistently outperform other methods with both 5\% and 10\% of training data, underscoring the superiority of our approach over strong baselines. These results indicate that our VFG-SSC is generalizable to different architectures, can be applied to various labeled settings, and applies to both outdoor and indoor scenarios.

\myheading{Qualitative results:} We present the qualitative results of our method, VFG-SSC, compared to baseline models on the SemanticKITTI and NYUv2 datasets in \cref{fig:SemKITTI_results} and \cref{fig:NYU_results}, respectively. In outdoor scenarios, VFG-SSC outperforms the baselines, particularly in accurately predicting various instance classes, such as cars and people. In contrast, other baselines mis-segment people and produce irregular car shapes (row 1). Additionally, VFG-SSC excels in capturing the broader scene layout, including long-range structures such as crossroads, sidewalks, and buildings (rows 2, 3). In indoor scenes, VFG-SSC effectively reconstructs the global scene layout, including both visible and occluded regions (row 2), while also recovering finer details of smaller objects, such as chairs, tables, and sofas (rows 1, 3).

\subsection{Ablation Study}

In this section, we ablate our approach on 5\% labeled data of SemanticKITTI with OccFormer as the SSC network $f_\theta$.

\myheading{Enhancement techniques:}  In \cref{tab:Fusion_Ablation}, we compare our attention-based alignment method against several enhancement techniques, including simple concatenation, BEVFusion  \cite{BEVFusion}, and OpenOccupancy's Adaptive Fusion \cite{OpenOccupancy}. Additionally, we include an attention-only baseline where 3D features and 3D clues are concatenated along the feature dimension, followed by Deformable Self-Attention \cite{zhu2020deformable}. We evaluate the quality of the pseudo-labels generated using training ground truth (GT) labels (in step 2) and assess the final model's predictions against validation GT labels. The results demonstrate that our method consistently outperforms the other techniques on both the training and validation splits. Notably, even with the additional clues from VFMs, simply integrating them into an SSC network does not result in significant performance improvements compared to baselines without VFMs. We attribute the substantial performance gain to our enhancement module, which effectively leverages these additional clues.

\begin{table}
    \centering


    \begin{tabular}{lcccc}
    \toprule
    \textbf{Enhancement}& \multicolumn{2}{c}{\textbf{Train Split}} & \multicolumn{2}{c}{\textbf{Val Split}}  \\
    \textbf{Methods} & \textbf{mIoU $\uparrow$} & \textbf{IoU $\uparrow$} & \textbf{mIoU $\uparrow$} & \textbf{IoU $\uparrow$} \\ 
    \midrule
    None & 8.37 & 31.60 & 9.48 & 33.12\\
    Concat & 13.03 & 42.29 & 10.01 & 34.05 \\
    BEVFusion & 12.80 & 42.15 & 10.00 & 34.72 \\
    OpenOcc & 12.42 & 40.08 & 10.37 & 34.37 \\
    Deformable & 13.84 & 37.40 & 10.04 & 34.30 \\
    Ours & \textbf{15.59} & \textbf{42.73} & \textbf{11.73} & \textbf{35.76} \\
    \bottomrule
    \end{tabular}
    \caption{Quantitative results of different fusion methods.}
    \label{tab:Fusion_Ablation}
\end{table}

\begin{table}[t]
    \centering
    \begin{tabular}{lcccc}
    \toprule
     \textbf{Depth} & \multicolumn{2}{c}{\textbf{Train Split}} & \multicolumn{2}{c}{\textbf{Val Split}}  \\
    \textbf{Model} & \textbf{mIoU $\uparrow$} & \textbf{IoU $\uparrow$} & \textbf{mIoU $\uparrow$} & \textbf{IoU $\uparrow$} \\
    \midrule
    Depth-Anything & 12.63 & 37.66 & 10.38 & 35.62 \\   
    Metric3Dv2 & 14.86 & 41.28 & 11.09 & \textbf{36.20} \\
    UniDepth & \textbf{15.59} & \textbf{42.73} & \textbf{11.73} & 35.76 \\
    \bottomrule
    \end{tabular}
    
    \caption{Study on different depth foundation models.}
    \label{tab:Depth_Ablation}
\end{table}

\myheading{Study on different depth estimation models:} 
We conducted experiments using various depth foundation models to generate 2D depth maps. Since the effectiveness of our method hinges on the accuracy of pseudo-labels, employing a more reliable depth predictor can potentially enhance the quality of the generated 3D clues. As shown in ~\cref{tab:Depth_Ablation}, our approach achieves the best results with UniDepth.  It's worth mentioning that advancements in depth estimation will further enhance our method's performance. In this experiment, we used SEEM \cite{SEEM} as the segmentation model.

\myheading{Study on different segmentation models:} 
We present a study on various semantic foundation models like SEEM \cite{SEEM} (pretrained on COCO), OpenSEED \cite{OpenSEED} (pretrained on COCO, Object365), and SegFormer \cite{xie2021segformer} (pretrained on CityScapes).
The comparison results reported in ~\cref{tab:Semantic_Ablation} reveal marginal distinctions among these aforementioned models. SegFormer obtains the best results as the domain between training (CityScapes) and testing (SemanticKITTI) data is not substantial for 2D segmentation. This suggests that our method consistently produces high-quality 3D cues across diverse segmentation foundation models. 

\begin{table}[t]
    \centering
    \begin{tabular}{lcccc}
    \toprule
   \textbf{Semantic} & \multicolumn{2}{c}{\textbf{Train Split}} & \multicolumn{2}{c}{\textbf{Val Split}}  \\
    \textbf{Model} & \textbf{mIoU $\uparrow$}  & \textbf{IoU $\uparrow$} & \textbf{mIoU $\uparrow$}  & \textbf{IoU $\uparrow$}\\ 
    \midrule

    OpenSEED & 14.80 & 40.90 & 11.40 & 36.68 \\
    SEEM & 15.59 & \textbf{42.73} & 11.73 & 35.76\\
    SegFormer & \textbf{16.18} & 41.74 & \textbf{11.94} & \textbf{36.78} \\
    \bottomrule
    \end{tabular}
    
    \caption{Study on different semantic foundation models.}
    \label{tab:Semantic_Ablation}
\end{table}
\myheading{Impact analysis of temporal and regularization:} Our analysis, detailed in ~\cref{tab:Components}, reveals several insights. In rows 1 and 2, we compare the performance of ST with and without the integration of a single frame of regularized 3D clues, noting a modest improvement of +0.09 mIoU. This enhancement can be attributed to the additional sparse information provided by the VFMs. In row 3, we evaluated the impact of incorporating temporal frames into 3D clue generation while omitting the regularization module, which resulted in a decrease of -0.62 mIoU. This drop indicates that the added temporal information from VFMs without regularization, introduces too much noise. However, when combining temporal accumulation with our regularization, we achieved a significant improvement of +2.25 mIoU. This demonstrates that while temporal information is valuable for VFG-SSC, regularization is essential for optimizing performance.

\begin{table}[th]
    \centering
    \setlength\tabcolsep{3pt} 

    \begin{tabular}{cccccc}
    \toprule
    & & \multicolumn{2}{c}{\textbf{Train Split}} & \multicolumn{2}{c}{\textbf{Val Split}} \\
    \textbf{Regularize} & \textbf{Temporal} & \textbf{mIoU $\uparrow$} & \textbf{IoU $\uparrow$} & \textbf{mIoU $\uparrow$} & \textbf{IoU $\uparrow$} \\
    \midrule
    & & 8.37 & 31.6 & 9.48 & 33.12\\
     \cmark & & 9.28 & 33.87 & 9.57 & 34.26\\
     & \cmark & 8.61 & 33.16 & 8.86 & 33.64\\
    \cmark & \cmark & \textbf{15.59} & \textbf{42.73} & \textbf{11.73} & \textbf{35.76}\\
     
    \bottomrule
    \end{tabular}
    \caption{Impact of VFG-SSC's temporal regularization.}
    \label{tab:Components}
\end{table}

\section{Conclusion}
\label{sec:conclusion}
In this paper, we have introduced a new approach for semi-supervised 3D semantic scene completion. Our method effectively utilizes 3D clues derived from 2D depth and segmentation foundation models and introduces an innovative fusion module to integrate these clues with 3D features, ensuring compatibility with most existing SSC techniques. Extensive experiments have demonstrated the effectiveness of our approach in both outdoor and indoor environments, achieving a minimal performance gap of just 15\% compared to fully supervised methods while using only 10\% of 3D semantic occupancy annotations. To the best of our knowledge, this is the first exploration of semi-supervised 3D SSC. 

\myheading{Limitations.} Given the reliance on using 3D clues from the 2D VFMs, our approach might not work well under challenging conditions (e.g., night-time, foggy, rainy) where VFMs fail to produce good results. 
Furthermore, large domain gaps between training and test data might harm our method, as our approach uses a small amount of labeled data to train.
\clearpage
\section*{Acknowledgements} We acknowledge Ho Chi Minh City University of Technology (HCMUT), VNU-HCM for supporting this study. 
\\ \\
Most of the work was done at VinAI Research.
\bibliography{aaai25}

\clearpage
\section{Supplementary Material}

In this supplementary material, we present additional experimental results that could not be illustrated in the main paper due to length limitations. In particular, we first provide implementation details of VFG-SSC for each backbone that our proposed framework can support. We then provide additional quantitative results, including class-wise IoU scores on SemanticKITTI and NYUv2. Next, we show qualitative and quantitative results for 3D clues generation and pseudo-labels generated in step on SemanticKITTI and NYUv2. We also provide more comparisons between recent self-supervised methods against our VFG-SSC on SemanticKITTI. Finally, additional qualitative results of our VFG-SSC on SemanticKITTI and NYUv2 validation and testing sets are demonstrated. 

\section{Additional Implementation Details}
\label{sec:addition_implementation}

\myheading{Backbone:} For all experiments with OccFormer \cite{OccFormer} and VoxFormer \cite{VoxFormer} on SemanticKITTI \cite{SemanticKITTI}, we use ConvNeXT-Base as the image backbone, we notice that this reduces the memory consumption compared to EfficientNetB7 and gives the models a slight boost in performance. For all experiments with MonoScene \cite{MonoScene} and OccDepth \cite{OccDepth}, we keep the EfficientNetB7 backbone since ConvNeXT-Base requires great effort to adapt to NYUv2.

\myheading{Data augmentation:} We list the data augmentations for each backbone in \cref{tab:data_aug}. Here we keep the augmentations the same as the original implementation of each backbone to ensure fair comparisons.

\myheading{Training semantic losses:} Details of the supervised loss for training each backbone are outlined in~\cref{tab:supervised_loss}.

\begin{table}[h]
\centering

\resizebox{\columnwidth}{!}{%
\begin{tabular}{ll}
\toprule
\textbf{Method}                     & \textbf{Supervised Losses}                                              \\
\midrule
OccFormer               & Binary cross-entropy (depth supervision), \\
& Mask classification\\
VoxFormer & Binary cross-entropy, Cross-entropy, \\
& Scene-class affinity                            \\ 
MonoScene & Cross-entropy, Scene-class affinity, \\
& Frustum proportion, Relation prior   \\
OccDepth & Binary cross-entropy, cross-entropy,  \\
& Scene-class affinity, Frustum proportion, \\
& Relation prior\\
\bottomrule
\end{tabular}%
}
\caption{Details of supervised losses for each backbone.}
\label{tab:supervised_loss}

\end{table}

\begin{table}[]
\centering
\setlength{\tabcolsep}{2pt}
\begin{tabular}{ll}
\toprule
\textbf{Method}                     & \textbf{Data Augmentations}                                              \\
\midrule
OccFormer                & Random color jittering, rotation, scaling, flipping  \\
VoxFormer & Random color jittering                            \\
MonoScene & Random color jittering and flipping                     \\
OccDepth & Random color jittering, gray scaling, \\ 
& rotation, scaling, flipping, erasing, and blurring
\\
\bottomrule
\end{tabular}%
\caption{Details of data augmentations for each backbone.}
\label{tab:data_aug}

\end{table}

\myheading{Other baselines:} For Self-Training \cite{Self-Training}, we use the supervised-only checkpoint to generate the pseudo-labels and then initialize the model from that checkpoint to continue training with the new pseudo-labels. For Mean-Teacher \cite{MeanTeacher}, we use the L2 loss and set the consistency weights to $10^3$ and the EMA rate to 0.99. For SelfOcc \cite{huang2023selfocc}, we apply a segmentation head similar to TPVFormer (cite TPVFormer) in place of the self-supervised SDF head and fine-tune with learning rate of $0.5\times 10^{-3}$.

\myheading{3D clues generation:} We provide the hyperparameter configurations for filtering in SemanticKITTI and NYUv2 as outlined in~\cref{tab:hyperparam_filtering}. It's important to highlight that our 3D clues necessitate spatial shapes consistent with the 3D features of the backbone. In SemanticKITTI, both OccFormer and VoxFormer models extract 3D features with dimensions of 128x128x16, thus we have adjusted the voxelization of our 3D clues to match this spatial setup. Similarly, for NYUv2, in alignment with MonoScene and OccDepth, we voxelized the 3D clues with dimensions of 60x36x60. A pseudo-code implementation of our regularization pipeline is presented in~\cref{lst:pseudo_code}.

\begin{listing*}[tb]%
\caption{Semantic and Geometric Temporal Regularization}%
\label{lst:pseudo_code}%
\begin{lstlisting}[language=Python]

def semantic_temporal_regularization(ref_num, poses, intrinsic, filter_params, ranges):
    '''
        Input:  
            ref_num (int): current frame to produce 3D clues
            poses (list[array]): list of 4x4 matrices with extrinsic information
            intrinsic (array): 3x3 array with camera parameters
            filter_params (dict): filtering parameters
            ranges (list): range bound of the voxels, with format [[x_min, x_max], [y_min, y_max], [z_min, z_max]]
        Output: 3D clues (HxWxD), where H,W,D is the size of the voxels
    '''
    ## Initalize Filters ##
    radius_filter = remove_radius_outlier(nb_points=filter_params['radius_filter']['nb_points'], radius=filter_params['radius_filter']['radius'])
    statistical_filter = remove_statistical_outlier(nb_neighbors=filter_params['statistic_filter']['nb_neighbors'], std_ratio=filter_params['statistic_filter']['std_ratio'])
    
    ## Load Target Point Clouds ##
    ref = poses[ref_num]
    targets = []
    min_num = max(ref_num-20, 0)
    max_num = min(ref_num+80, len(poses))
    for i in range(min_num, max_num, 5): 
        targets.append(i)
    
    point_clouds = []
    semantics = []
    for i in targets:
        target = poses[i]
        target2ref = np.matmul(inv(ref), target) 
        
        depth = read_depth(i)        
        point_cloud = unproject(depth, intrinsic, target2ref)
        point_clouds.append(point_cloud)

        semantic = read_semantics(i)
        semantics.append(semantic.reshape(-1))

    ## Transform and Accumulate to Reference ##
    all_pc = []
    all_semantic = []
    for i in range(len(point_clouds)):
        point_cloud = point_clouds[i]
        semantic = semantics[i]
        
        classes = np.unique(semantic)
        for cl in classes:
            if cl == 255:
                continue
            cl_mask = (semantic == cl)
            point_cloud_class = point_cloud[cl_mask]
            
            radius_ind = radius_filter(point_cloud_class)
            statis_ind = statistical_filter(point_cloud_class[radius_ind])

            all_pc.append(point_cloud_class[radius_ind][statis_ind])
            all_semantic.append(np.full(point_cloud_class[radius_ind][statis_ind].shape, cl)
            
    all_pc_np = np.concatenate(all_pc, axis=0) #Nx3
    all_semantic_np = np.concatenate(all_semantic, axis=0) #Nx1
    clues_3D = voxelize_xyz_and_sem(all_pc_np, all_semantic_np, ranges) #HxWxD
    return clues_3d
    \end{lstlisting}
\end{listing*}

\begin{table}[t]
\centering
\footnotesize
\setlength{\tabcolsep}{5pt}
\begin{tabular}{lll}
\toprule
\textbf{Dataset}  & \textbf{Filters} &\textbf{Hyper-parameters}\\
\midrule
\multirow{2}{*}{SemanticKITTI} & Radius & nb\_points=8, radius=0.1 \\
& Statistical & nb\_neighbors=10, std\_ratio=1.5 \\
\midrule
\multirow{2}{*}{NYUv2} & Radius & nb\_points=50, radius=0.08 \\
& Statistical & nb\_neighbors=30, std\_ratio=1.5 \\
\bottomrule
\end{tabular}%
\caption{Details of hyperparameters for filtering.}
\label{tab:hyperparam_filtering}

\end{table}

\begin{figure*}
    \centering
    \includegraphics[width=\linewidth]{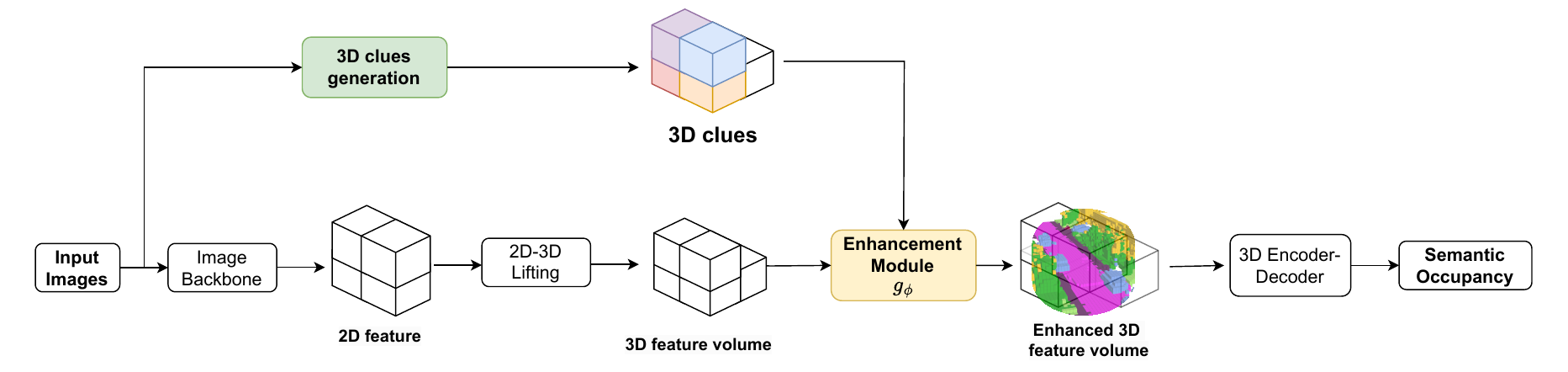}
    \caption{Detailed VFG-SSC pipeline for OccFormer, MonoScene, OccDepth}
    \label{fig:short-a}
\end{figure*}

\myheading{Enhancement Module:} In this section, we show detailed pipelines for all supported backbones. 
For backbones utilizing 2D-3D lifting mechanisms like Lift-Splat-Shoot (OccFormer~\cite{OccFormer}) or parameter-free interpolation (MonoScene~\cite{MonoScene}, OccDepth~\cite{OccDepth}), an initial 3D volume feature is generated using the respective mechanism. Next, we fuse this lifted 3D volume feature with 3D clues to obtain refined 3D features to use in subsequent modules. 
This process is depicted in \cref{fig:short-a}.

\begin{figure*}
    \centering
    \includegraphics[width=\linewidth]{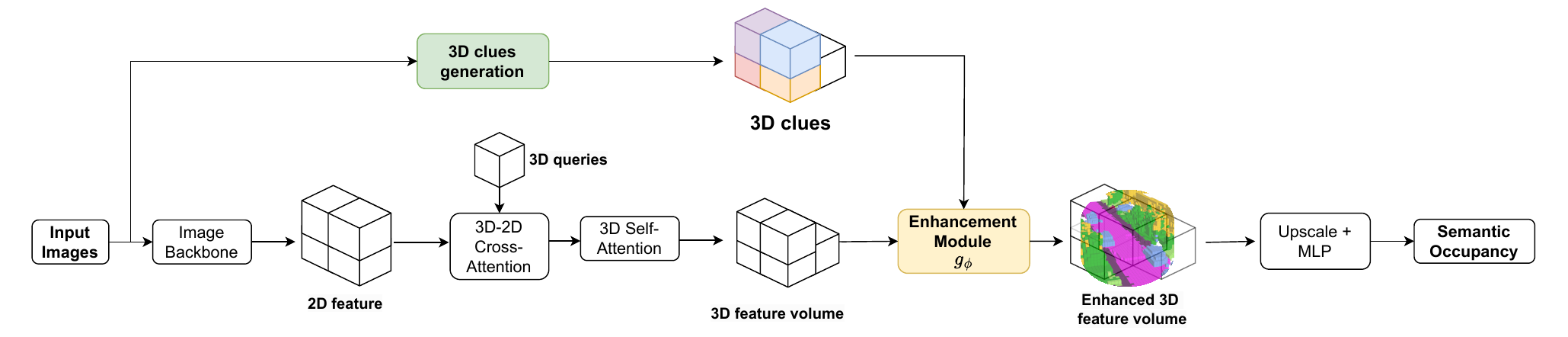}
    \caption{Detailed VFG-SSC pipeline for VoxFormer}
    \label{fig:short-b}
\end{figure*}

On the other hand, for backbones employing 3D-2D transformers such as VoxFormer~\cite{VoxFormer}, a set of queried positions in 3D is projected back into the 2D feature space to create an initial 3D volume feature. This initial feature is then refined through 3D self-attention layers. Post self-attention, we fuse the 3D feature with 3D clues to refine it further. Finally, upscaling and MLP layers decode the refined 3D feature into semantic occupancy. This process is illustrated in~\cref{fig:short-b}.



\section{Class-wise Results}
\label{sec:addition_quantitative}
\cref{tab:SemanticKITTIvalset,tab:NYUtestset} display class-wise IoU scores for various semi-supervised baselines on the SemanticKITTI validation set and the NYUv2 test set, respectively. Our Semi-SSC demonstrates significant performance gain, surpassing others by a large margin of +2 mIoU with the VoxFormer/OccFormer backbones. 

\section{Scaling with unlabeled data}
In the main paper, we presented experiments in the \textbf{partial labeling} scenario, where we label only 1/5/10\% of the data. Another practical aspect of semi-supervised is the \textbf{scaling with unlabeled data} scenario, where we supplement the fully labeled SemanticKITTI dataset with an additional 200\% of unlabeled data (around 6,800 images) from the KITTI dataset \cite{Geiger2012CVPR}, which shares the same camera setup as SemanticKITTI. In~\cref{tab:scaling}, we compare our approach to fully supervised training using 100\% of the labeled SemanticKITTI data.

The results show that simply adding unlabeled images with pseudo labels does not necessarily improve the performance of SSC models. However, by applying our method, we can further enhance the performance of SSC models.

\begin{table}[]
\small
\centering
\setlength{\tabcolsep}{1pt}
\begin{tabular}{lccc}
\toprule
 \textbf{SSC Network}  & \textbf{\% labeled:unlabeled} & \textbf{mIoU $\uparrow$} & \textbf{IoU $\uparrow$} \\
\midrule
ST (Occ) & 100 : 200 & 13.31 (-0.77) & 36.74 (+0.70) \\
ST (Vox) & 100 : 200 & 13.86 (-0.54) & 43.66 (+0.16) \\
VFG-SSC (Occ) & 100 : 200 & 14.86 (+0.78) & 38.69 (+2.65) \\
VFG-SSC (Vox) & 100 : 200 & 14.88 (+0.40) & 44.00 (+0.50) \\

\bottomrule
\end{tabular}
\vspace{-5pt}
\caption{Results of scaling with unlabeled data.}
\label{tab:scaling}
\end{table}

\section{Qualitative Results of 3D Clues Generation}
\label{sec:3D_clues_compare}
\begin{figure*}[tb]
  \centering
  \includegraphics[width=\textwidth]{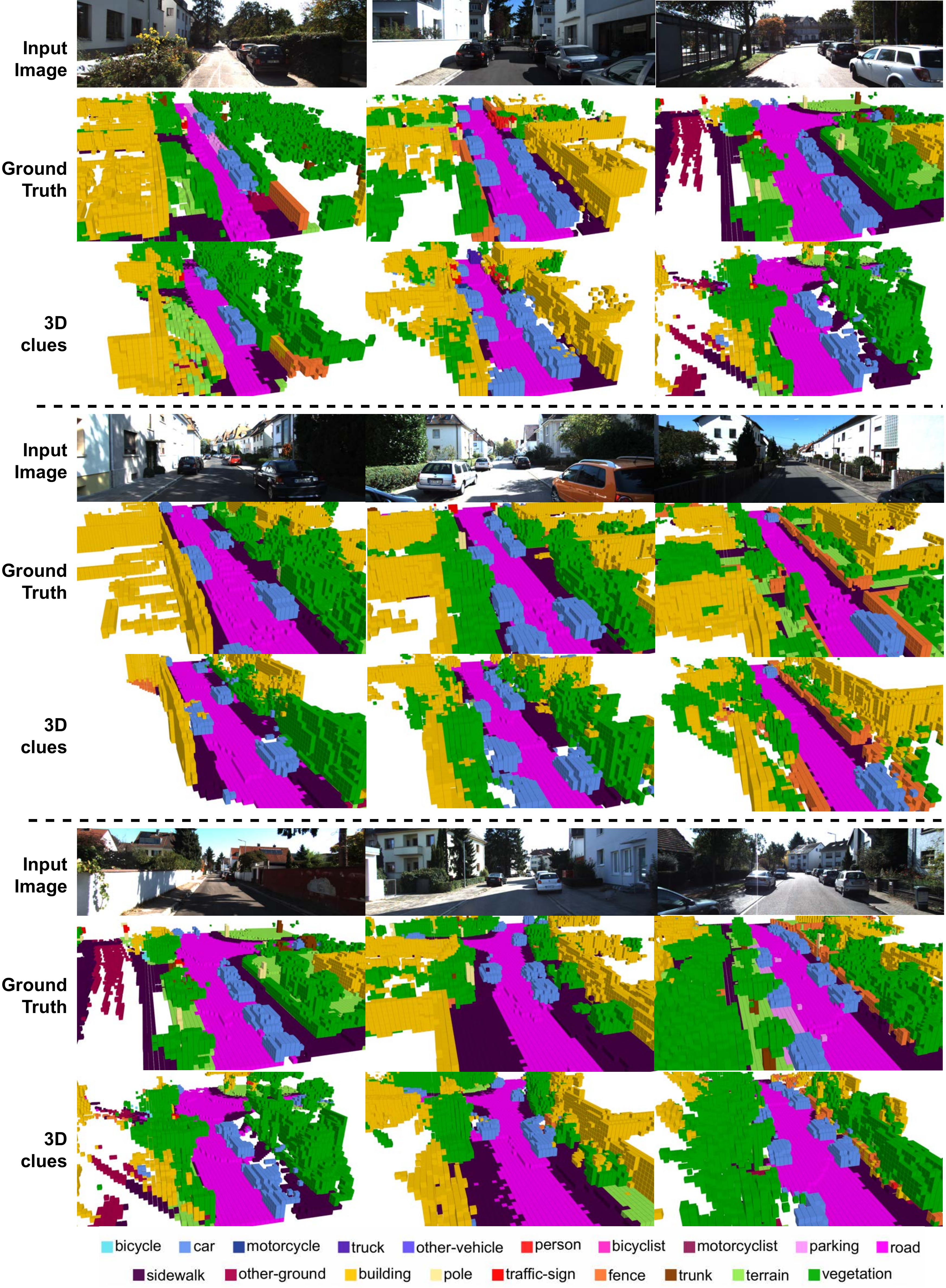}
  \caption{Qualitative results for 3D clues generation on SemanticKITTI\cite{SemanticKITTI}. Our
3D clues can capture the detailed reconstruction of the scene.}
  \label{fig:3dclues_compare}
\end{figure*}

\begin{figure*}[tb]
  \centering
  \includegraphics[width=\textwidth]{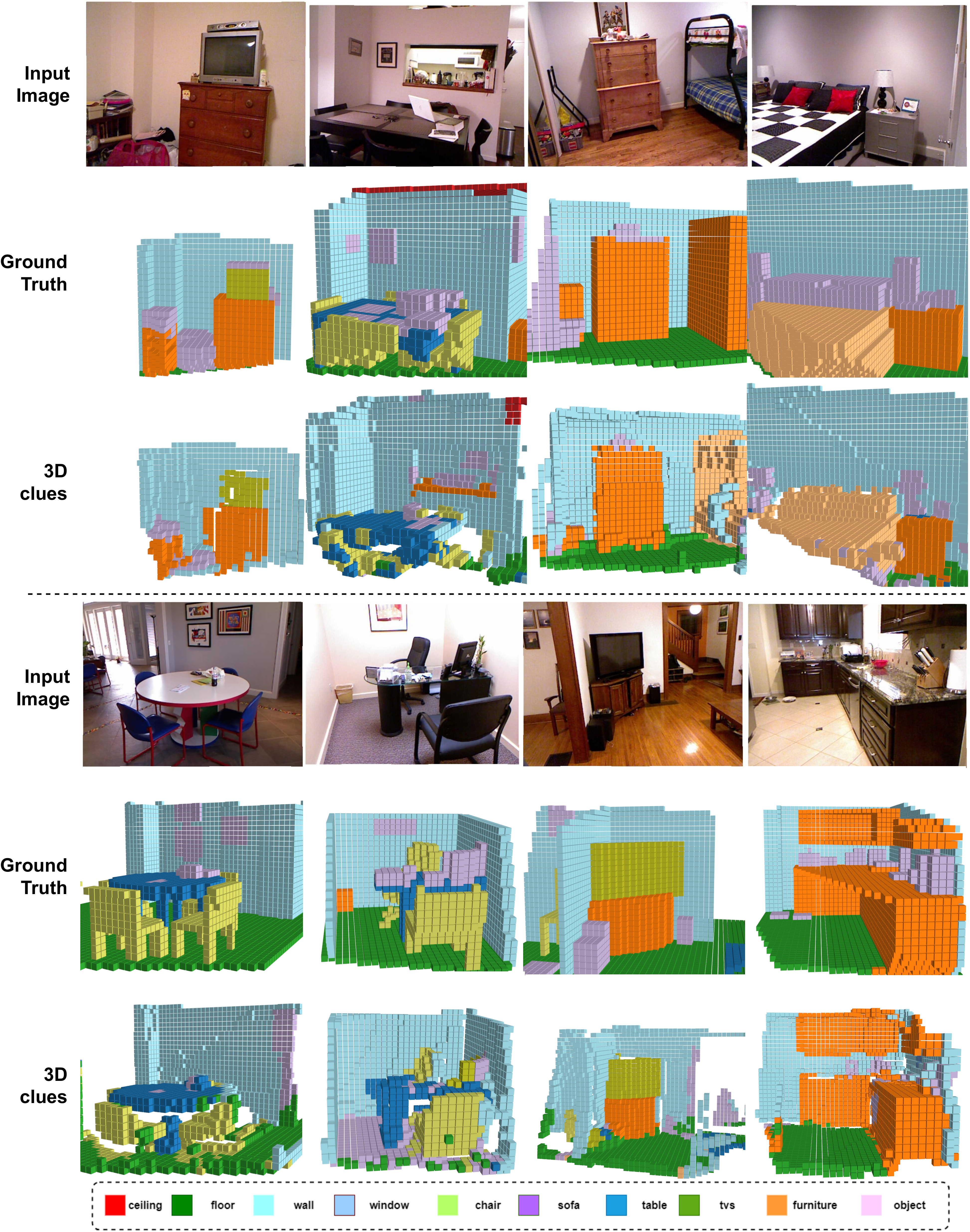}
  \caption{Qualitative results for 3D clues generation on NYUv2\cite{silberman2012indoor}. Our
3D clues capture detailed reconstruction of the scene and lead to better results in low-data scenarios.}
  \label{fig:nyu_3d_clues_compare}
\end{figure*}

To showcase our filtering's effectiveness in reconstructing 3D structures, we present additional results on SemanticKITTI (Figure \ref{fig:3dclues_compare}) and NYUv2 (Figure \ref{fig:nyu_3d_clues_compare}). We use OccFormer for SemanticKITTI and MonoScene for NYUv2. Despite scale ambiguity, our method consistently aligns semantics with ground-truth labels, demonstrating its ability to generate 3D clues and address data scarcity.

\section{Qualitative Results of Pseudo-Labels (Step 2)}
\label{sec:pseudo_label_compare}

\begin{figure*}[tb]
  \centering
  \includegraphics[width=\textwidth]{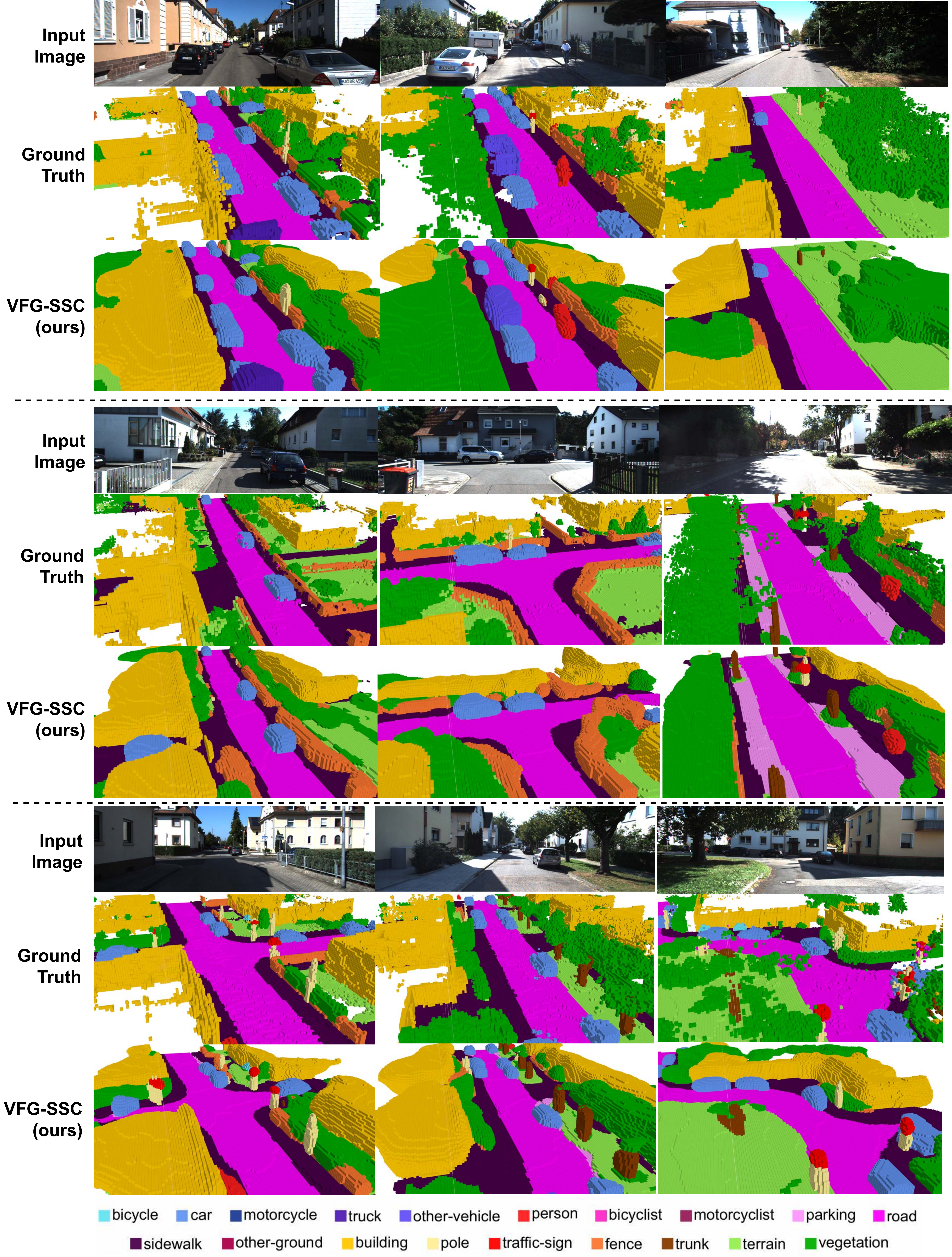}
  \caption{Qualitative results for pseudo-labels (step 2) on SemanticKITTI\cite{SemanticKITTI}.}
  \label{fig:pseudo_semantic}
\end{figure*}

\begin{figure*}[tb]
  \centering
  \includegraphics[width=\textwidth]{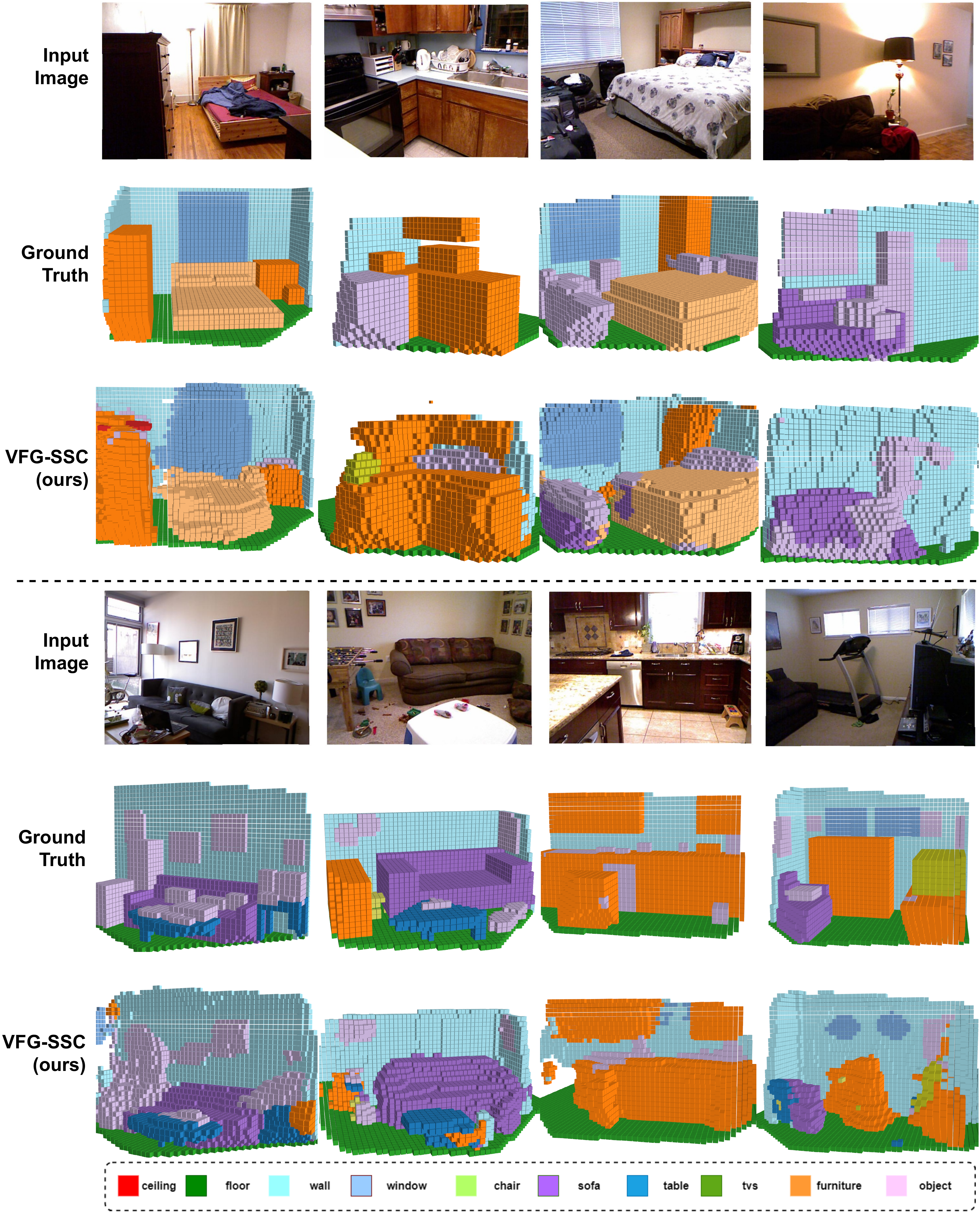}
  \caption{Qualitative results for pseudo-labels (step 2) on NYUv2\cite{silberman2012indoor}.}
  \label{fig:nyu_pseudo_semantic}
\end{figure*}

Here we provide more insight into step 2 of our framework by plotting pseudo labels from SemanticKITTI (\cref{fig:pseudo_semantic}) and NYUv2 (\cref{fig:nyu_pseudo_semantic}). We choose OccFormer and MonoScene as our backbone for SemanticKITTI and NYUv2, respectively. As can be seen, our pseudo-labels show great intra-class consistency and have a sharper boundary between different semantic classes.

\begin{table*}[t]
\small
\centering
\setlength{\tabcolsep}{1.8pt}

\begin{tabular}{lllccccccccccccccccccccc}
\toprule
\textbf{Split} & \textbf{Model} & \textbf{Method} &
  \textbf{mIoU$\uparrow$ } &
  \textbf{IoU$\uparrow$ } &
  \rotatebox[origin=c]{90}{\textbf{road}} &
  \rotatebox[origin=c]{90}{\textbf{sidewalk}} &
  \rotatebox[origin=c]{90}{\textbf{parking}} &
  \rotatebox[origin=c]{90}{\textbf{other-grnd}} &
  \rotatebox[origin=c]{90}{\textbf{building}} &
  \rotatebox[origin=c]{90}{\textbf{car}} &
  \rotatebox[origin=c]{90}{\textbf{truck}} &
  \rotatebox[origin=c]{90}{\textbf{bicycle}} &
  \rotatebox[origin=c]{90}{\textbf{motorcycle}} &
  \rotatebox[origin=c]{90}{\textbf{other-vehi}} &
  \rotatebox[origin=c]{90}{\textbf{vegetation}} &
  \rotatebox[origin=c]{90}{\textbf{trunk}} &
  \rotatebox[origin=c]{90}{\textbf{terrain}} &
  \rotatebox[origin=c]{90}{\textbf{person}} &
  \rotatebox[origin=c]{90}{\textbf{bicyclist}} &
  \rotatebox[origin=c]{90}{\textbf{motorcyclist}} &
  \rotatebox[origin=c]{90}{\textbf{fence}} &
  \rotatebox[origin=c]{90}{\textbf{pole}} &
  \rotatebox[origin=c]{90}{\textbf{traffic-sign}} \\ 
\midrule
& SelfOcc & Finetuning & 
     5.5 & 
    29.6 & 
    43.2 & 
    15.5 & 
    0.0 &
    0.0 & 
    6.6 & 
    18.6 &
    0.0 &
    0.0 &
    0.1 & 
    1.3 &
    11.2 &
    0.2 &
    14.0 &
    0.0 &
    0.0 &
    0.0 & 
    1.3 &
    1.2 &
    0.1
    \\
\cmidrule{2-24}
\multirow{9}{*}{1\%} &\multirow{5}{*}{OccFormer}& Supervised-only &
  6.6 &
  26.3 &
  45.5 &
  18.9 &
  0.0 &
  0.0 &
  6.5 &
  17.5 &
  0.0 &
  0.0 &
  0.9 &
  4.7 &
  11.3 &
  0.0 &
  18.4 &
  0.0 &
  0.0 &
  0.0 &
  1.2 &
  0.2 &
  0.0 \\
  && Mean-Teacher &
  6.8 &
  26.5 &
  44.4 &
  20.3 &
  0.0 &
  0.0 &
  5.9 &
  16.3 &
  0.0 &
  0.0 &
  \textbf{3.3} &
  5.2 &
  11.8 &
  0.0 &
  19.9 &
  0.0 &
  0.0 &
  0.0 &
  1.1 &
  0.0 &
  0.0 \\
 && Self-Training&
  6.7 &
  26.6 &
  47.1 &
  19.7 &
  0.0 &
  0.0 &
  7.0 &
  17.1 &
  0.0 &
  0.0 &
  0.7 &
  4.5 &
  11.2 &
  0.0 &
  19.4 &
  0.0 &
  0.0 &
  0.0 &
  0.9 &
  0.5 &
  0.0 \\
  && VFG-SSC  &
  \textbf{9.4} & 
  \textbf{30.4} & 
  \textbf{56.4} & 
  \textbf{27.3} & 
  0.0 &
  0.0 &
  \textbf{12.5} &
  \textbf{21.1} &
  0.0 & 
  0.0 & 
  1.7 &
  \textbf{8.1} & 
  \textbf{13.5} & 
  \textbf{0.1} &
  \textbf{29.6} & 
  0.0 & 
  0.0 & 
  0.0 & 
  \textbf{4.5} & 
  \textbf{2.8} & 
  \textbf{1.1}
  \\
  \cmidrule{2-24}
&\multirow{4}{*}{VoxFormer}& Supervised-only &
  6.3 &
  34.0 &
  38.9 &
  17.2 &
  0.0 &
  0.0 &
  10.6 &
  16.4 &
  0.0 &
  0.0 &
  0.0 &
  0.1 &
  19 &
  0.0 &
  15.0 &
  0.0 &
  0.0 &
  0.0 &
  2.0 &
  1.3 &
  0.2 \\
  && Mean-Teacher  &
  7.2 &
  \textbf{39.0} &
  44.9 &
  16.8 &
  0.0 &
  0.0 &
  11.1 &
  18.6 &
  0.0 &
  0.0 &
  0.8 &
  0.0 &
  \textbf{20.7} &
  0.2 &
  19.4 &
  0.0 &
  0.0 &
  0.0 &
  1.8 &
  1.7 &
  0.8 \\
 && Self-Training  &
  7.6 &
  35.4 &
  44.9 &
  20.8 &
  0.0 &
  0.0 &
  9.9 &
  17.1 &
  0.0 &
  \textbf{0.2} &
  1.0 &
  1.6 &
  19 &
  \textbf{1.6} &
  21.9 &
  0.0 &
  0.0 &
  0.0 &
  2.5 &
  3.0 &
  0.8 \\
  && VFG-SSC  &
  \textbf{9.3} & 
  36.8 & 
  \textbf{50.1} &
  \textbf{25.3} & 
  0.0 & 
  0.0 & 
  \textbf{12.6} & 
  \textbf{23.5} & 
  0.0 & 
  0.0 & 
  \textbf{1.8} & 
  \textbf{7.8} & 
  20.2 & 
  0.0 & 
  \textbf{24.5} & 
  0.0 & 
  0.0 & 
  0.0 & 
  \textbf{5.8} & 
  \textbf{4.0} &
  \textbf{1.4}
  \\
  \midrule  
& SelfOcc & Finetuning & 
  7.5 & 
  32.4 & 
  45.9 &
  19.2 & 
  1.3 &
  0.3 & 
  9.7 & 
  21.2 & 
  0.1 & 
  0.0 & 
  0.3 &
  3.6 &
  14.9 &
  0.8 &
  17.1 &
  0.0 &
  0.0 &
  0.0 &
  1.9 & 
  1.5 &
  0.1 
  \\
\cmidrule{2-24}
\multirow{9}{*}{5\%} &\multirow{5}{*}{OccFormer}& Supervised-only &
  9.2 &
  32.7 &
  49.7 &
  22.0 &
  8.5 &
  0.0 &
  11.3 &
  20.9 &
  1.5 &
  0.4 &
  1.0 &
  9.0 &
  17.1 &
  2.3 &
  24.3 &
  1.2 &
  0.0 &
  0.0 &
  1.6 &
  3.1 &
  0.0 \\
  && Mean-Teacher &
  9.4 &
  31.7 &
  50.3 &
  23.3 &
  6.2 &
  0.0 &
  11.7 &
  20.0 &
  3.0 &
  0.0 &
  \textbf{2.9} &
  \textbf{10.6} &
  18.0 &
  2.3 &
  23.4 &
  0.4 &
  0.0 &
  0.0 &
  2.7 &
  \textbf{3.4} &
  0.0 \\
 && Self-Training  &
  9.5 &
  33.1 &
  50.0 &
  21.6 &
  9.7 &
  0.0 &
  11.2 &
  20.2 &
  \textbf{8.9} &
  \textbf{0.8} &
  0.7 &
  6.1 &
  17.3 &
  2.7 &
  25.4 &
  0.0 &
  0.0 &
  0.0 &
  1.9 &
  3.3 &
  0.0 \\

  && VFG-SSC  &
    \textbf{11.7} & 
    \textbf{35.8} & 
    \textbf{56.9} & 
    \textbf{30.0} & 
    \textbf{14.5} & 
    0.0 & 
    \textbf{15.3} & 
    \textbf{22.3} & 
    6.2 & 
    0.0 & 
    1.7 & 
    9.4 & 
    \textbf{19.6} & 
    \textbf{3.4} & 
    \textbf{31.7} & 
    \textbf{3.0} & 
    0.0 &
    0.0 & 
    \textbf{5.6} & 
    3.1 & 
    0.0
    \\
  \cmidrule{2-24}
&\multirow{4}{*}{VoxFormer}& Supervised-only &
  9.6 &
  39.7 &
  48.7 &
  19.1 &
  3.4 &
  0.0 &
  15.3 &
  22 &
  0.2 &
  0.1 &
  2.3 &
  3.2 &
  23.3 &
  4.9 &
  24.9 &
  0.6 &
  0.0 &
  0.0 &
  5.0 &
  6.1 &
  2.5 \\
  && Mean-Teacher  &
  9.6 &
  37.7 &
  46.6 &
  22.5 &
  6.4 &
  0.0 &
  15.7 &
  21.8 &
  1.4 &
  0.3 &
  2.0 &
  3.4 &
  22.9 &
  4.5 &
  18.6 &
  1.0 &
  0.0 &
  0.0 &
  5.3 &
  6.1 &
  3.3 \\
 && Self-Training&
  10.5 &
  39.8 &
  49.3 &
  25.4 &
  \textbf{6.8} &
  0.0 &
  13.4 &
  21.7 &
  0.5 &
  1.0 &
  \textbf{3.5} &
  4.1 &
  23.4 &
  \textbf{5.5} &
  \textbf{27.7} &
  \textbf{1.8} &
  0.0 &
  0.0 &
  5.5 &
  5.3 &
  \textbf{4.5} \\
  && VFG-SSC  &
   \textbf{11.1} & 
   \textbf{40.0} & 
   \textbf{51.7} & 
   \textbf{26.5} & 
   2.9 &
   \textbf{0.1} & 
   \textbf{16.6} & 
   \textbf{23.6} & 
   \textbf{8.3} & 
   \textbf{1.6} & 
   1.3 & 
   \textbf{5.1} & 
   \textbf{24.1} & 
   5.3 & 
   27.1 & 
   1.2 & 
   0.0 & 
   0.0 & 
  \textbf{6.5} & 
   \textbf{6.6} & 
   3.3\\
  \midrule  
& SelfOcc & Finetuning &
 8.3 & 33.7 &
 48.6 & 
 20.3 &
 6.3 & 
 0.0 &
 10.9 & 
 21.8 & 
 0.1 & 
 0.0 & 
 0.1 & 
 1.7 & 
 16.1 &
 1.2 &
 18.8 & 
 0.3 &
 0.0 &
 0.0 &
 2.2 & 
 1.5 & 
 0.1
 \\
\cmidrule{2-24}
\multirow{9}{*}{10\%} & \multirow{5}{*}{OccFormer}& Supervised-only &
  10.4 &
  34.2 &
  54.4 &
  23.6 &
  13.9 &
  0.0 &
  12.1 &
  22.0 &
  0.6 &
  2.1 &
  2.0 &
  \textbf{9.5} &
  17.6 &
  3.4 &
  29.2 &
  0.3 &
  0.3 &
  0.0 &
  2.7 &
  3.1 &
  1.4 \\
  && Mean-Teacher  &
  10.5 &
  33.3 &
  54.5 &
  26.1 &
  14.5 &
  0.0 &
  12.5 &
  22.1 &
  0.0 &
  0.0 &
  \textbf{2.8} &
  8.4 &
  17.5 &
  3.2 &
  29.3 &
  2.1 &
  0.0 &
  0.0 &
  2.9 &
  2.9 &
  1.0 \\
 && Self-Training  &
  10.5 &
  34.0 &
  54.0 &
  22.7 &
  13.7 &
  0.0 &
  12.2 &
  20.0 &
  2.7 &
  2.5 &
  0.9 &
  8.1 &
  \textbf{18.2} &
  3.6 &
  29.8 &
  1.3 &
  \textbf{2.9} &
  0.0 &
  3.1 &
  2.7 &
  1.1 \\
  && VFG-SSC  &
    \textbf{12.4} & \textbf{35.2} &
    \textbf{58.6} & 
    \textbf{28.8} & 
    \textbf{16.5} & 
    0.0 & 
   \textbf{14.7} & 
    \textbf{22.5} & 
    \textbf{5.6} & 
    \textbf{2.6} & 
    2.5 & 
    9.4 & 
    17.9 &
    \textbf{3.8} & 
    \textbf{32.2} & 
    \textbf{3.6} & 
    2.0 & 
    0.0 & 
    \textbf{7.3} & 
    \textbf{4.6} & 
    \textbf{2.6} \\
  \cmidrule{2-24}
&\multirow{4}{*}{VoxFormer}& Supervised-only &
  10.5 &
  40.3 &
  48.9 &
  22.3 &
  10.8 &
  \textbf{0.1} &
  16.7 &
  22.5 &
  0.8 &
  1.7 &
  0.9 &
  3.1 &
  23.4 &
  5.3 &
  27.1 &
  1.0 &
  0.0 &
  0.0 &
  5.5 &
  5.5 &
  4.3 \\
  && Mean-Teacher  &
  10.6 &
  38.7 &
  48.1 &
  22.4 &
  10.1 &
  0.0 &
  16.8 &
  22.9 &
  4.9 &
  0.9 &
  1.9 &
  2.4 &
  23.3 &
  5.4 &
  27.1 &
  0.4 &
  0.0 &
  0.0 &
  4.0 &
  5.8 &
  \textbf{4.1} \\
 && Self-Training  &
  11.1 &
  41.0 &
  51.0 &
  25.9 &
  9.2 &
  0.0 &
  17.0 &
  22.3 &
  0.2 &
  2.1 &
  \textbf{2.6} &
  2.8 &
  24.0 &
  \textbf{6.0} &
  \textbf{28.7} &
  1.7 &
  \textbf{0.3} &
  0.0 &
  5.9 &
  6.6 &
  3.8 \\
  & & VFG-SSC  &
  \textbf{12.2} & 
  \textbf{41.6} & 
  \textbf{53.1} & 
  \textbf{26.4} & 
  \textbf{9.8} & 
  0.0 & 
  \textbf{18.7} & 
 \textbf{ 25.3 }& 
  \textbf{10.6} & 
  2.1 & 
  2.0 & 
  \textbf{5.0} & 
  \textbf{25.1} & 
  5.4 & 
  28.3 & 
  \textbf{1.8} & 
  0.0 & 
  0.0 & 
  \textbf{6.8} & 
  \textbf{7.5} & 
  3.6
  \\
\bottomrule
\end{tabular}%
\caption{Comparisons between our method and other semi-supervised methods on the validation set of the SemanticKITTI~\cite{SemanticKITTI} dataset. The best results are highlighted in \textbf{bold}.}
\label{tab:SemanticKITTIvalset}

\end{table*}

\begin{table*}[tb]
\centering
\small
\setlength{\tabcolsep}{4pt}

\begin{tabular}{lllccccccccccccc}
\toprule
\textbf{Split} & \textbf{Model} & \textbf{Method} &
  \textbf{mIoU$\uparrow$ } &
  \textbf{IoU$\uparrow$ } &
  \rotatebox[origin=c]{90}{\textbf{ceiling}} &
  \rotatebox[origin=c]{90}{\textbf{floor}} &
  \rotatebox[origin=c]{90}{\textbf{wall}} &
  \rotatebox[origin=c]{90}{\textbf{window}} &
  \rotatebox[origin=c]{90}{\textbf{chair}} &
  \rotatebox[origin=c]{90}{\textbf{bed}} &
  \rotatebox[origin=c]{90}{\textbf{sofa}} &
  \rotatebox[origin=c]{90}{\textbf{table}} &
  \rotatebox[origin=c]{90}{\textbf{tvs}} &
  \rotatebox[origin=c]{90}{\textbf{furniture}} &
  \rotatebox[origin=c]{90}{\textbf{objects}}  \\ 
\midrule
\multirow{6}{*}{5\%} &\multirow{3}{*}{MonoScene}& Supervised-only &
  13.4 &
  29.2 &
  0.1 &
  93.0 &
  3.8 &
  3.8 &
  3.3 &
  19.3 &
  3.9 &
  1.4 &
  0.3 &
  13.9 &
  4.1  \\
  && Self-Training &
  14.0 &
  30.6 &
  0.3 &
  93.0 &
  4.5 &
  3.9 &
  \textbf{4.4} &
  19.4 &
  4.4 &
  2.4 &
  0.2 &
  16.1 &
  4.2 \\
  && VFG-SSC  &
  \textbf{18.2 }&
  \textbf{43.3} &
  \textbf{0.7} &
  93.0 &
  \textbf{16.8} &
  \textbf{7.1} &
  4.2 &
  \textbf{24.9} &
  \textbf{12.0} &
  \textbf{9.5} &
  \textbf{1.1} &
  \textbf{21.5} &
  \textbf{9.7} \\
  \cmidrule{2-16} 
  &\multirow{3}{*}{OccDepth}& Supervised-only &
  16.7 &
  31.0 &
  0.4 &
  92.3 &
  7.3 &
  3.3 &
  5.6 &
  25.9 &
  13.9 &
  7.3 &
  2.0 &
  17.8 &
  7.7  \\
  && Self-Training &
  16.7 &
  30.5 &
  0.4 &
  92.4 &
  6.6 &
  3.0 &
  4.9 &
  27.3 &
  15.4 &
  6.9 &
  2.1 &
  18.0 &
  6.7 \\
  && VFG-SSC  &
  \textbf{22.5} &
  \textbf{43.6 }&
  \textbf{1.1 }&
  \textbf{92.5} &
  \textbf{15.9} &
  \textbf{10.3 }&
  \textbf{10.4} &
  \textbf{37.3} &
  \textbf{24.8} &
  \textbf{12.7} &
  \textbf{6.1} &
  \textbf{26.1} &
  \textbf{10.8} \\
\midrule
\multirow{6}{*}{10\%} &\multirow{3}{*}{MonoScene}& Supervised-only &
  17.2 &
  34.7 &
  0.4 &
  92.5 &
  6.5 &
  4.9 &
  6.3 &
  24.5 &
  20.6 &
  8.6 &
  0.8 &
  18.7 &
  5.0  \\
  && Self-Training&
  17.3 &
  35.2 &
  0.7 &
  92.4 &
  6.9 &
  4.7 &
  6.1 &
  19.0 &
  21.0 &
  9.5 &
  0.4 &
  19.4 &
  6.4 \\
  && VFG-SSC  &
  \textbf{22.2} &
  \textbf{43.4} &
  \textbf{8.4} &
  \textbf{93.0} &
  \textbf{16.6} &
  \textbf{11.2} &
  \textbf{12.4} &
  \textbf{28.8} &
  \textbf{24.2} &
  \textbf{13.2} &
  \textbf{2.9} &
  \textbf{24.7} &
  \textbf{9.3} \\
  \cmidrule{2-16} 
  &\multirow{3}{*}{OccDepth}& Supervised-only &
  19.0 &
  37.8 &
  2.4 &
  92.4 &
  10.3 &
  5.0 &
  8.8 &
  30.9 &
  20.1 &
  9.4 &
  0.2 &
  19.9 &
  9.7  \\
  && Self-Training&
  19.4 &
  38.1 &
  1.8 &
  \textbf{92.5} &
  9.7 &
  6.0 &
  9.5 &
  32.4 &
  21.8 &
  10.1 &
  0.2 &
  20.3 &
  9.2 \\
  && VFG-SSC  &
  \textbf{23.9} &
  \textbf{44.8}&
  \textbf{4.6} &
  92.3 &
  \textbf{15.8} &
  \textbf{9.4} &
  \textbf{11.7} &
  \textbf{42.2} &
  \textbf{30.7} &
  \textbf{14.5} &
  \textbf{2.6} &
  \textbf{27.8} &
  \textbf{11.4} \\
\bottomrule
\end{tabular}%
\caption{Quantitative comparisons between our method and other semi-supervised methods on the testing set of the NYUv2 \cite{silberman2012indoor} dataset. The best results are highlighted in \textbf{bold}.}
\label{tab:NYUtestset}

\end{table*}

\end{document}